%% file: main.tex
\documentclass[10pt,twocolumn,letterpaper]{article}

\usepackage{cvpr}              %

\input{preamble}

\definecolor{cvprblue}{rgb}{0.21,0.49,0.74}
\usepackage{graphicx}
\usepackage{booktabs}
\usepackage{multicol}
\usepackage{multirow}
\usepackage{arydshln}
\usepackage[pagebackref,breaklinks,colorlinks,citecolor=cvprblue]{hyperref}

\title{ZeroGrasp: Zero-Shot Shape Reconstruction Enabled Robotic Grasping}

\DeclareMathAlphabet\mathbfcal{OMS}{cmsy}{b}{n}

\begin{document}

\author{
Shun Iwase$^{1,2}$\and
Muhammad Zubair Irshad$^{2}$\and
Katherine Liu$^{2}$\and
Vitor Guizilini$^{2}$\and
Robert Lee$^{3}$\and
Takuya Ikeda$^{3}$\and
Ayako Amma$^{3}$\and
Koichi Nishiwaki$^{3}$\and
Kris Kitani$^{1}$\and
Rareș Ambruș$^{2}$\and
Sergey Zakharov$^{2}$\and
$^1$Carnegie Mellon University \qquad $^2$Toyota Research Institute \qquad $^3$Woven by Toyota
}

\maketitle

\begin{abstract}
Robotic grasping is a cornerstone capability of embodied systems. Many methods directly output grasps from partial information without modeling the geometry of the scene, leading to suboptimal motion and even collisions.
To address these issues, we introduce ZeroGrasp, a novel framework that simultaneously performs 3D reconstruction and grasp pose prediction in near real-time.
A key insight of our method is that occlusion reasoning and modeling the spatial relationships between objects is beneficial for both accurate reconstruction and grasping.
We couple our method with a novel large-scale synthetic dataset, which comprises $1$M photo-realistic images, high-resolution 3D reconstructions and $11.3$B physically-valid grasp pose annotations for $12$K objects from the Objaverse-LVIS dataset. 
We evaluate ZeroGrasp on the GraspNet-1B benchmark as well as through real-world robot experiments. ZeroGrasp achieves state-of-the-art performance and generalizes to novel real-world objects by leveraging synthetic data. \href{https://sh8.io/#/zerograsp}{https://sh8.io/\#/zerograsp}

\end{abstract}

\begin{figure}
    \centering
    \includegraphics[width=\linewidth]{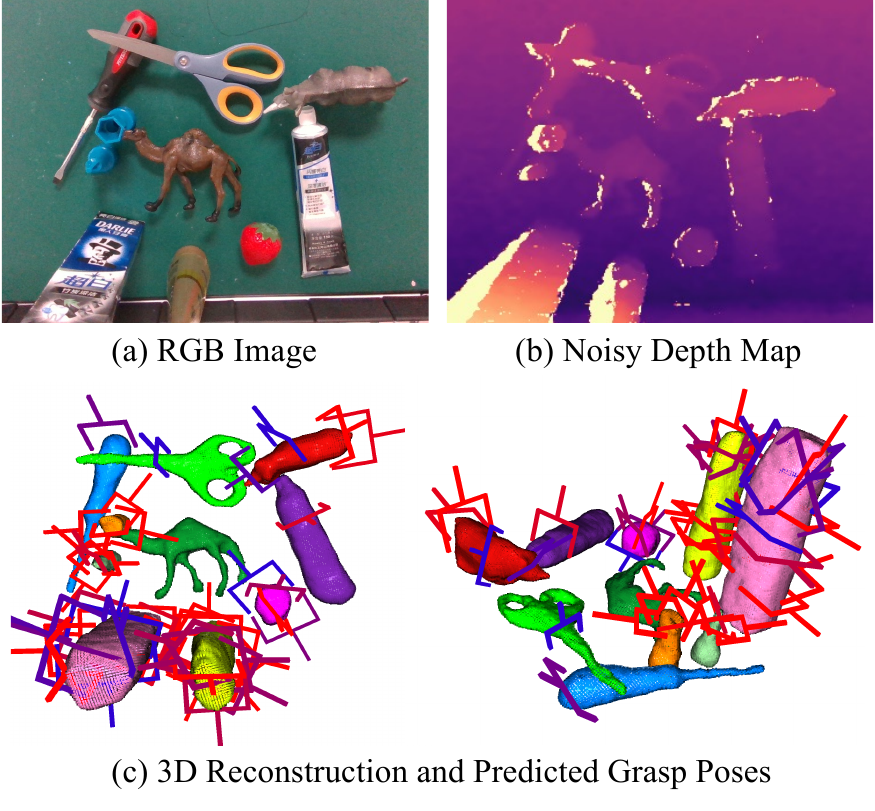}
    \caption{ZeroGrasp simultaneously reconstructs objects at high-resolution and predicts grasp poses from a single RGB-D image in near real-time ($5$FPS).}
    \label{fig:teaser}
\end{figure}

\begin{figure*}[t]
    \vspace{-1.2cm}
    \centering \scalebox{0.95}{
    \includegraphics[width=\linewidth]{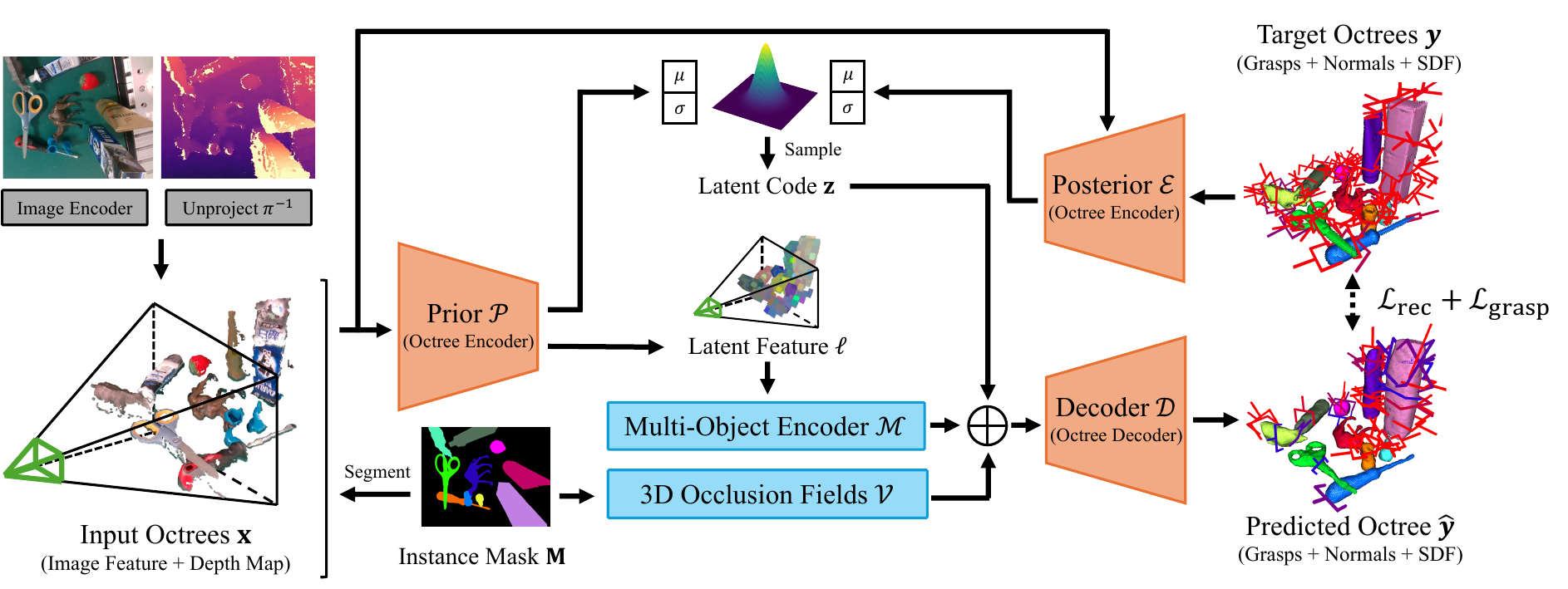}}
    \caption{Overview of ZeroGrasp, a novel method for simultaneous 3D reconstruction and 6D grasp pose predictions from a single-view RGB-D image. The input octree $\mathbf{x}$ is first fed into the octree-based CVAE (components with orange boxes). The multi-object encoder takes its latent feature $\ell$ to learn multi-object reasoning at the latent space. Further, 3D occlusion fields encode inter- and self-occlusion information via simple ray casting. The output features from the multi-object encoder and 3D occlusion fields are concatenated with the latent code $\mathbf{z}$, and 3D shapes and grasp poses are predicted by the decoder.} 
    \label{fig:overview}
\end{figure*}

\section{Introduction}
Safe and robust robotic grasping requires accurate geometric understanding of target objects, as well as their surroundings. However, most previous grasp detection methods~\cite{fang2020graspnet,jian2021,Ma_2024_cvpr,liang2019pointnetgpd,morrison2018closing,mousavian2019graspnet} do not explicitly model the geometry of the target objects, which can lead to unexpected collisions and unstable contact with target objects.
Although several methods~\cite{shen2023F3RM,Ma_2024_cvpr} leverage multi-view images to reconstruct the target objects in advance, this process introduces additional computational overhead and requires a more complex setup. Multi-view reconstruction is also often impractical for objects placed within confined spaces like shelves or boxes. Furthermore, the lack of large-scale datasets with ground-truth 3D shapes and grasp pose annotations further complicates accurate 3D reconstruction from a single RGB-D image.
Recently, several works~\cite{choy20194d,10.1145/3072959.3073608,Iwase_ECCV_2024} demonstrate that sparse voxel representations outperform volumetric and NeRF-like implicit shape representations in terms of runtime, accuracy, and resolution, particularly for regression-based zero-shot 3D reconstruction.

To leverage reconstruction methods using sparse voxel representations for robotic grasping, it is essential to develop new approaches that can reason about both within a unified framework. To this end, we propose ZeroGrasp, a novel framework for near real-time 3D reconstruction and 6D grasp pose prediction. Our key hypothesis is that improved 3D reconstruction quality enhances grasp pose prediction, specifically by leveraging physics-based contact constraints and collision detection, which are essential for accurate grasping. Since robotic environments often involve multiple objects with inter-object occlusions and close contacts, ZeroGrasp introduces two key components: a multi-object encoder and 3D occlusion fields. These components effectively model inter-object relationships and occlusions, thus, improving reconstruction quality. In addition, we design a simple refinement algorithm to improve grasp poses using the predicted reconstruction. Because the reconstruction is highly accurate, it provides reliable contact points and collision masks between the gripper and the target object, which we use to refine the grasp poses.

In addition to our proposed model, we also create a real-world dataset for evaluation, the ReOcS dataset, and a synthetic datasets for training, the ZeroGrasp-11B dataset. 
The ReOcS dataset is a real-world evaluation dataset of 3D reconstruction, with three splits representing different degrees of occlusion. We use this dataset to assess robustness to occlusions. The ZeroGrasp-11B dataset is a large-scale synthetic dataset designed to train models with zero-shot robotic grasping capability, containing high-quality and diverse 3D models from Objaverse-LVIS dataset~\cite{objaverse}, as shown in \Cref{tab:dataset}.

We evaluate both the baseline and our methods, showing that our approach --- trained on the GraspNet-1B dataset~\cite{fang2020graspnet} alone, as well as on a combination of the GraspNet-1B dataset and ZeroGrasp-11B --- achieves state-of-the-art performance on the GraspNet-1B benchmark. Our ablation studies further show that the proposed components enhance both reconstruction and grasp pose prediction quality. Finally, we conduct real-robot evaluations to demonstrate its generalizability in real-world scenarios.

Our contributions are summarized as follows:
\begin{itemize}
    \item We propose ZeroGrasp, a novel framework for simultaneous 3D reconstruction and 6D grasp pose prediction using an octree-based conditional variational autoencoder (CVAE). ZeroGrasp achieves the best performance on the GraspNet-1B benchmark and real-robot evaluation.
    \item We introduce a multi-object encoder and 3D occlusion fields to model inter-object relationships and occlusions. 
    \item We propose a simple grasp pose refinement algorithm that improves grasp accuracy using the reconstructed 3D shape.
    \item We create two datasets, 1) the ReOcS dataset for evaluating 3D reconstruction under occlusions, and 2) the ZeroGrasp-11B dataset for zero-shot robotic grasping.
\end{itemize}

\begin{table*}[t!]
    \caption{\textbf{Dataset comparisons.} We create a large-scale grasp detection dataset for zero-shot robotic grasping using $12$K 3D models from Objaverse-LVIS dataset~\cite{objaverse}. Our ZeroGrasp-11B dataset includes $1$ million RGB-D images and physics-based dense 6D grasp annotations of cluttered scenes.}
    \centering
    \setlength{\tabcolsep}{8pt}
    \scalebox{0.78}{
    \begin{tabular}{l|ccccccccc}
       Dataset & \# Images & \# 3D Models & \# Grasps & \# Cat. & Type & Modality & Grasp Alg. & Grasp Rep. \\ \hline
       Cornel~\cite{jian2021} & $1$K & $0.2$K & $8$K & 16 & Real & RGB-D & Manual & Planar \\
       Jacquard~\cite{Depierre2018JacquardAL} & $54$K & $11$K & $1.1$M & N/A & Sim. & RGB-D & Physics & Planar \\
       Zhang \etal~\cite{10.1109/IROS40897.2019.8967869} & $4.7$K & $\mathbf{{\approx}15}$\textbf{K} & $100$K & N/A & Real & RGB & Manual & Planar \\ \hline
       VR-Grasping-101~\cite{10.1109/ICRA.2018.8460609} & $10$K & $0.1$K & $4.8$M & 7 & Sim. & RGB-D & Manual & 6D \\
       GraspNet-1Billion~\cite{fang2020graspnet} & $97$K & $0.1$K & \underline{1.2B} & 30-35 & Real & RGB-D & Analytical & 6D \\
       ACRONYM~\cite{acronym2020} & N/A & $9$K & $17.7$M & \underline{262} & Sim. & N/A & Physics & 6D \\
       REGRAD~\cite{9681218} & \underline{900K} & \textbf{50K} & $100$M & 55 & Sim. & N/A & Physics & 6D \\
       HouseCat6D~\cite{jung2022housecat6d} & $23.5$K & $0.2$K & $10$M & 10 & Real & RGB-D+P & Physics & 6D \\
       Grasp-Anything-6D~\cite{nguyen2024language} & $\mathbf{1}$\textbf{M} & N/A & $200$M & N/A & Synth. & RGB + ZoeDepth~\cite{https://doi.org/10.48550/arxiv.2302.12288} & Analytical & 6D \\
       ZeroGrasp-11B (Ours) & $\mathbf{1}$\textbf{M} & \underline{12K} & $\mathbf{11.3}$\textbf{B} & \textbf{606} & Sim. & RGB-D & Physics & 6D \\
    \end{tabular}}
    \label{tab:dataset}
\end{table*}

\section{Related Works}

\paragraph{Regression-based 3D reconstruction.}
Regression-based 3D reconstruction from a single-view RGB-D image~\cite{huang2023zeroshape,ren2024xcube,OccupancyNetworks,Peng2020ECCV,irshad2022centersnap,irshad2022shapo,bozic2021transformerfusion,dai2018scancomplete,10.1007/978-3-031-19824-3_30,choy20194d,lunayach2023fsd,huang2023nksr,li2023voxformer,heppert2023carto,wu2023multiview,Boulch_2022_CVPR,shen2021dmtet,Liu2023MeshDiffusion,Park_2019_CVPR,yu2021pointr,yan2022shapeformer,autosdf2022,cheng2023sdfusion,varley2017shape,PSWang2020,song2016ssc,zhang2022cgca,10160350,Zhang_2019_ICCV} have been a major focus of research in 3D computer vision.
These methods explore different 3D representations, including dense voxel grids~\cite{Peng2020ECCV,Li2020aicnet,yan2022shapeformer,li2023voxformer}, sparse voxel grids~\cite{10.1145/3072959.3073608,choy20194d,williams2024fvdb} (\eg octree~\cite{10.1145/3072959.3073608}, VDB~\cite{williams2024fvdb}, hash table~\cite{choy20194d}, and etc.), and implicit representations~\cite{yu2021pointr,huang2023zeroshape,Boulch_2022_CVPR,wu2023multiview}. Nevertheless, dense voxel grid and implicit representations face limitations in output resolution due to expensive memory and computational costs. On the other hand, several works~\cite{huang2023zeroshape,ren2024xcube,williams2024fvdb,10.1145/3072959.3073608} show that sparse voxel representations such as an octree and VDB~\cite{10.1145/2487228.2487235} enable high-resolution 3D reconstruction with faster runtime thanks to its efficient hierarchical structure. 
Alternatively, single-view reconstruction through novel view synthesis achieves impressive results. Recent works such as GeNVS~\cite{chan2023genvs}, Zero-1-to-3~\cite{liu2023zero1to3}, 3DiM~\cite{Watson2022NovelVS}, and InstantMesh~\cite{xu2024instantmesh} leverage diffusion models to render multi-view images given a canonical camera pose. However, these approaches are slow (often over $10$ seconds) and inter-object occlusions degrade the performance significantly. Further, integrating grasp pose prediction is nontrivial. Thus, we adopt an octree as a shape representation and design our framework based on octree-based U-Net~\cite{10.1145/3072959.3073608}.

\paragraph{Regression-based Grasp Pose Prediction.}
Traditional grasp pose prediction methods typically assume prior knowledge of 3D objects and often rely on simplified analytical models based on force closure principles~\cite{1087483,844081}. Recently, tremendous progress has been made in learning-based approaches~\cite{fang2020graspnet,Wang_2021_ICCV,mahler2017dex,mousavian2019graspnet} which have allowed models to predict 6D grasp poses directly from RGB(-D) images and point clouds. This has enabled the regression of grasp poses in highly cluttered scenes without explicitly modeling object geometries. However, this could result in unstable grasping and unexpected collision, as accurately learning collision avoidance and precise contact points remains challenging. Although some methods~\cite{varley2017shape,chisari2024centergrasp,jiang2021synergies} explore 3D reconstruction to improve grasp prediction, their choices of shape representations and network architectures often limit its full potential.

\paragraph{Zero-shot robotic grasping.}
Zero-shot robotic grasping refers to the ability to grasp unseen target objects without prior knowledge. To achieve this, there are mainly two directions --- (1) optimizing grasp poses at test time based on contact points using reconstructed or ground-truth 3D shapes~\cite{grady2021contactopt,Ma_2024_cvpr}, and (2) augmenting or synthesizing large-scale grasp data to improve generalization~\cite{acronym2020,fang2020graspnet,morrison2020egad}. For instance, Ma \etal~\cite{Ma_2024_cvpr} propose a contact-based optimization algorithm to refine initial grasp poses by using a reconstructed 3D scene from multi-view RGB-D images. Existing large-scale grasp pose datasets such as ACRONYM~\cite{acronym2020}, GraspNet-1B~\cite{fang2020graspnet}, and EGAD~\cite{morrison2020egad} explore the second approach. Nevertheless, they are limited to object diversity or missing annotations like RGB-D images. Inspired by these two approaches, we aim to improve generalization to unseen objects with a simple and efficient grasp pose refinement algorithm that utilizes predicted reconstructions. Further, we create a large-scale synthetic dataset for grasp pose detection. Our dataset comprises high-quality and diverse objects, as well as $1$M photorealistic RGB images and physics-based grasp pose annotations.

\section{Proposed Method}
Our goal is to build an efficient and generalizable model for simultaneous 3D shape reconstruction and grasp pose prediction from a single RGB-D observation, and to demonstrate that the predicted reconstructions can be used to refine grasp poses via contact-based constraints and collision detection. In this section, we describe the network architecture and grasp pose refinement algorithm.

\paragraph{3D shape representation.}
We adopt an octree as a shape representation where attributes such as image features, the signed distance function (SDF), normals, and grasp poses are defined at the deepest level of the octree. For instance, we represent an input octree as a tuple of voxel centers $\mathbf{p}$ at the final depth, associated with image features $\mathbf{f}$,
\begin{equation}
  \mathbf{x} = \left(\mathbf{p}, \mathbf{f} \right), \,\, \mathbf{p} \in \mathbb{R}^{N \times 3}, \mathbf{f} \in \mathbb{R}^{N \times D}.
\end{equation}
where $N$ is the number of voxels.
Unlike point clouds, an octree structure enables efficient depth-first search and recursive subdivision to octants, making it ideal for high-resolution shape reconstruction and dense grasp pose prediction in a memory and computationally efficient manner.

\paragraph{Grasp pose representation.}
We represent grasp poses using a general two-finger parallel gripper model, as used in GraspNet~\cite{fang2020graspnet}.
Specifically, our grasp poses consist of the following components: view graspness score $\mathbf{s} \in \mathbb{R}^{M}$, which indicates the robustness of grasp positions~\cite{Wang_2021_ICCV}; quality $\mathbf{q} \in \mathbb{R}^{M}$, computed using the force closure algorithm~\cite{1087483}; view direction $\mathbf{v} \in \mathbb{R}^{3M}$; angle $\mathbf{a} \in \mathbb{R}^{M}$; width $\mathbf{w} \in \mathbb{R}^{M}$; and depth $\mathbf{d} \in \mathbb{R}^{M}$:
\begin{equation}
  \boldsymbol{g} = \begin{bmatrix} \mathbf{s} & \mathbf{q} & \mathbf{v} & \mathbf{a} & \mathbf{w} & \mathbf{d} \end{bmatrix},
\end{equation}
where $M$ denotes the number of total grasps in the target octree, and the closest grasp poses within a 5 mm radius are assigned to each point. If none exists, we set its corresponding graspness to $0$. In GraspNet-1B and ZeroGrasp-11B datasets, each point is annotated with a dense set of grasp poses covering all combinations of views, angles, and depths ($300 \times 12 \times 4$).
With the grasp poses $\mathbf{g}$, the target octree is defined as 
\begin{equation}
    \mathbf{y} = \left(\mathbf{p}^{gt}, \mathbf{f}^{gt} \right) = \left(\mathbf{p}^{gt}, \begin{bmatrix}
      \boldsymbol{\phi} & \mathbf{n} & \mathbf{g}
    \end{bmatrix}\right),
\end{equation}
where $\boldsymbol{\phi} \in \mathbb{R}^{M}$ is the SDF, and $\mathbf{n} \in \mathbb{R}^{M\times3}$ is normal vectors of the target octree.

\subsection{Architecture}
Given input octrees $\mathbf{x}$, composed of per-instance partial point clouds derived from depth maps and instance masks, along with their corresponding image features, we aim to predict 3D reconstructions and grasp poses $\mathbf{\hat{y}}$ represented as octrees. ZeroGrasp is built upon an octree-based U-Net~\cite{10.1145/3072959.3073608} and conditional variational autoencoder (CVAE)~\cite{rempe2021humor} to model shape reconstruction uncertainty and grasp pose prediction, while maintaining near real-time inference. We present two key innovations to improve its accuracy and generalization. Specifically, we introduce (1) \textbf{multi-object encoder} to model spatial relations between objects via a 3D transformer in the latent space, enabling collision-free 3D reconstructions and grasp poses, and (2) \textbf{3D occlusion fields}, a novel 3D occlusion representation which captures inter-object occlusions to enhance shape reconstruction in occluded regions. %

\begin{figure}[t]
    \centering
    \scalebox{0.9}{
    \includegraphics[width=\linewidth]{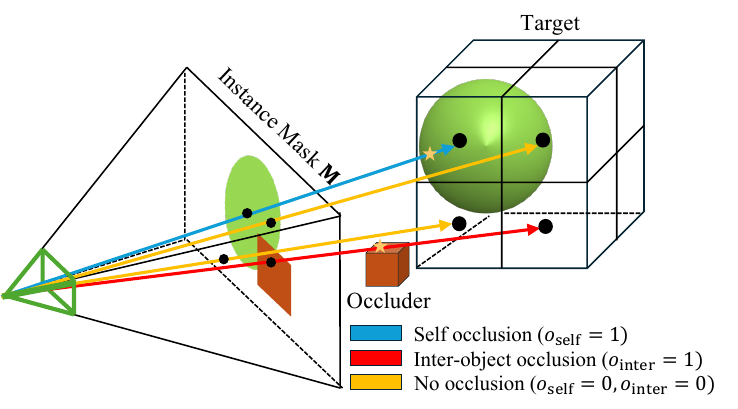}}
    \caption{\textbf{3D occlusion fields} localize occlusion information by casting rays from the camera to the voxel centers around the target object and performing depth tests. Specifically, if a ray intersects the target object, a self-occlusion flag $o_{\text{self}}$ is set to 1. If it intersects non-target objects, an inter-object occlusion flag $o_{\text{inter}}$ is set to 1.}
    \label{fig:3d_occlusion_fields}
\end{figure}

\paragraph{Octree feature extraction.}
An RGB image $\mathbf{I} \in \mathbb{R}^{H \times W \times 3}$ is encoded to extract an image feature $\mathbf{W}$. We fine-tune SAM 2~\cite{ravi2024sam2} to generate 2D instance masks $\mathbf{M} \in \mathbb{R}^{H \times W}$ and $\mathbf{M}_i$ represents an $i$-th object mask 
The image feature map is then unprojected into 3D space by $\left(\mathbf{q}_i,\mathbf{w}_i\right) = \pi^{-1}\left(\mathbf{W}, \mathbf{D}, \mathbf{K}, \mathbf{M}_i \right)$ where $\mathbf{q}_i$ and $\mathbf{w}_i$ denote 3D point cloud and its corresponding features of an $i$-th object, respectively. Here, $\pi$ is the unprojection function, $\mathbf{D} \in \mathbb{R}^{H \times W}$ is the depth map and $\mathbf{K} \in \mathbb{R}^{3 \times 3}$ denotes the camera intrinsics. The 3D point cloud features are converted to an octree $\mathbf{x}_i = \left(\mathbf{p}_i, \mathbf{f}_i \right) =  \mathcal{G}\left(\mathbf{q}_i, \mathbf{w}_i\right)$ where $\mathcal{G}$ is the conversion function from the point cloud and its features to an octree. 

\begin{figure*}[t]
  \begin{minipage}{0.6\linewidth}
    \centering
    \vspace{-0.5cm}
    \includegraphics[width=\linewidth]{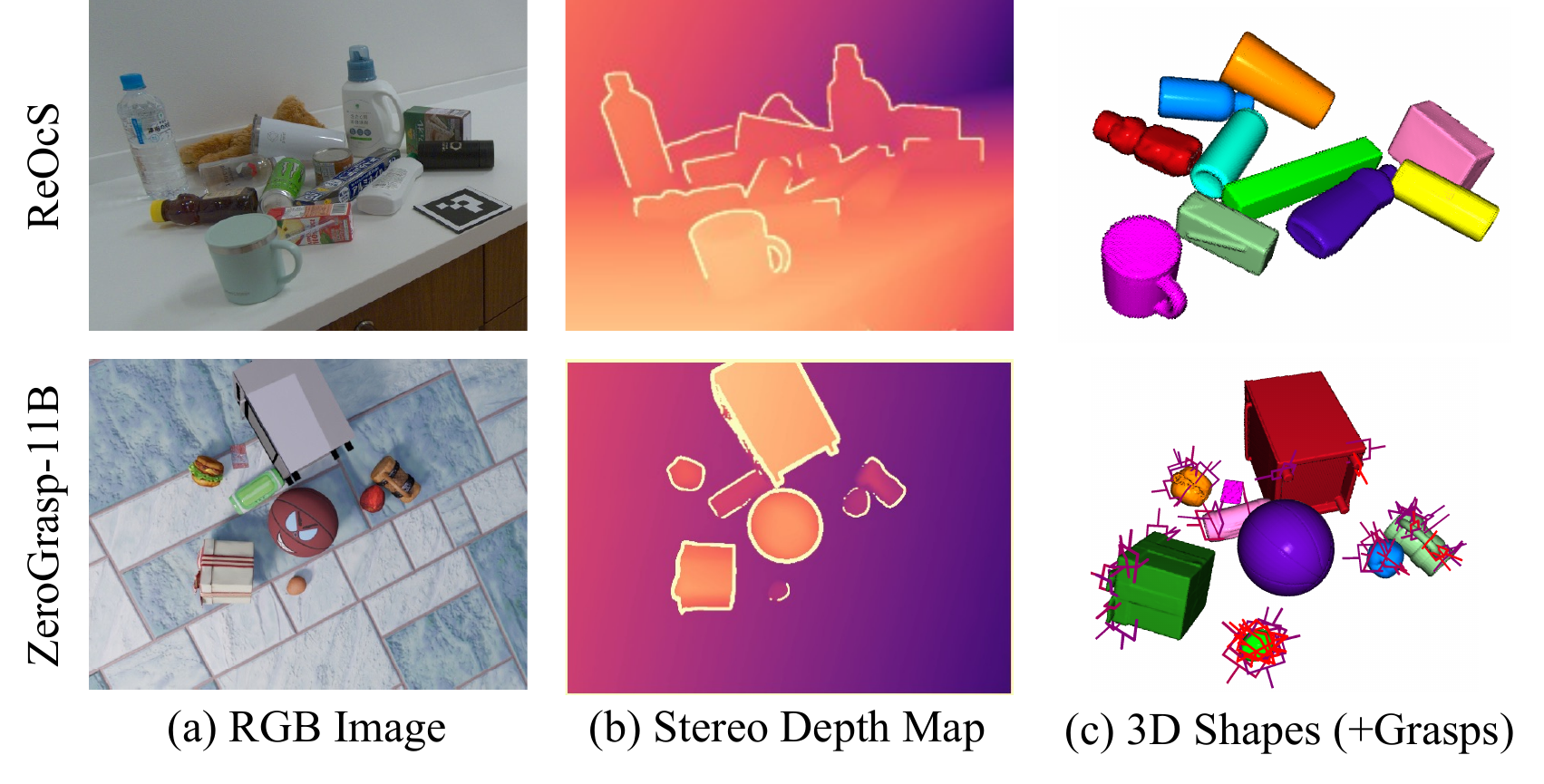}
    \vspace{-0.8cm}
    \caption{Example RGB images, stereo depth maps, 3D shapes and grasp poses from the ReOcs and ZeroGrasp-11B datasets. The grasp poses of the ZeroGrasp-11B dataset are subsampled by grasp-NMS~\cite{fang2020graspnet} for better visibility of the 3D shapes and grasps. More examples are found in the supplementary material.}
    \label{fig:dataset}
  \end{minipage}\hspace{.5cm}
  \begin{minipage}{0.37\linewidth}
    \centering
    \scalebox{0.7}{
    \includegraphics[width=\linewidth]{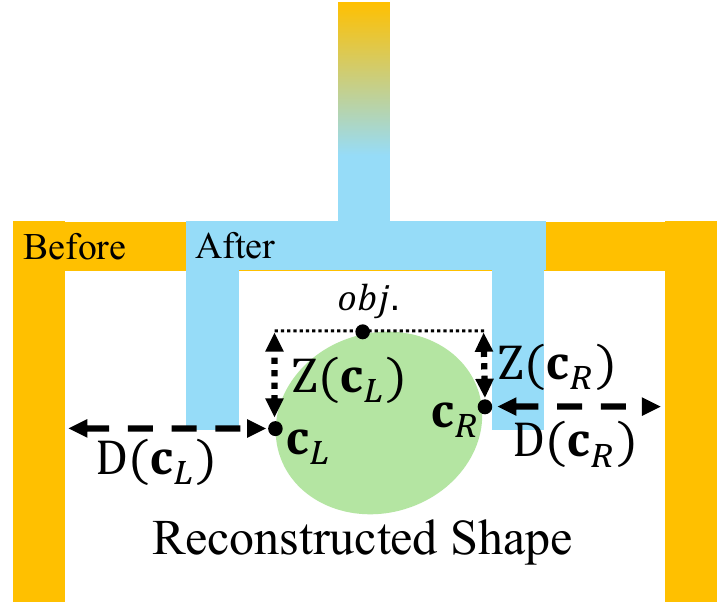}}
    \caption{\textbf{Contact-based constraints} are used to effectively refine grasp poses. We first obtain contact points $c_L$ and $c_R$. Next, the contact distance $D\left(c_{L|R}\right)$, and the depth is computed by $Z\left(c_{L|R}\right)$ are computed. Finally, the width and height of the grasp is refined based on \cref{eq:update_width} and \cref{eq:update_depth}.}
    \label{fig:collision_constraints}
  \end{minipage}
  \vspace{-0.3cm}
\end{figure*}

\paragraph{Octree-based CVAE.}
 To improve the shape reconstruction quality, ZeroGrasp utilizes probabilistic modeling through an octree-based conditional variational autoencoder (CVAE) to address the inherent uncertainty in single-view shape reconstruction, which is crucial for improving both reconstruction and grasp pose prediction quality. Inspired by \cite{rempe2021humor}, our Octree-based CVAE consists of an encoder $\mathbfcal{E}$, prior $\mathbfcal{P}$, and decoder $\mathbfcal{D}$ to learn latent representations of 3D shapes and grasp poses together as diagonal Gaussian. Concretely, the encoder $\mathbfcal{E}\left(\mathbf{z}_i \mid \mathbf{x}_i, \mathbf{y}_i\right)$ learns to predict the latent code $\mathbf{z}_i$ based on the predicted and ground-truth octrees $\mathbf{x}_i$ and $\mathbf{y}_i$. 
The prior $\mathbfcal{P}\left({\boldsymbol\ell}_i, \mathbf{z}_i \mid \mathbf{x}_i\right)$ takes the octree $\mathbf{x}_i$ as input and computes the latent feature ${\boldsymbol\ell}_i \in \mathbb{R}^{N^{\prime}_i \times D^{\prime}}$ and code $\mathbf{z}_i \in \mathbb{R}^{D^{\prime}}$ where $N^{\prime}_i$ and $D^{\prime}$ are the number of points and the dimension of the latent feature.
Internally, the latent code is sampled from predicted mean and variance via a reparameterization trick~\cite{Kingma2013AutoEncodingVB}. The decoder $\mathbfcal{D}\left(\mathbf{y}_i \mid {\boldsymbol\ell}_i, \mathbf{z}_i, \mathbf{x}_i\right)$ predicts a 3D reconstruction along with grasp poses. To save computational cost, the decoder predicts occupancy at each depth, discarding grid cells with a probability below $0.5$. Only in the final layer does the decoder predict the SDF, normal vectors and grasp poses as well as occupancy. During training, KL divergence between the encoder and prior is minimized such that their distributions are matched.

\paragraph{Multi-object encoder.}
The prior $\mathbfcal{P}$ computes features per object, lacking the capability of modeling global spatial arrangements for collision-free reconstruction and grasp pose prediction. To address this, we incorporate a transformer in the latent space, composed of $K$ standard Transformer blocks with self-attention and RoPE~\cite{su2021roformer} positional encoding, following in~\cite{Iwase_ECCV_2024}. The multi-object encoder $\mathcal{M}$ takes voxel centers $\mathbf{r}_{i} \in \mathbb{R}^{N^{\prime}_i \times 3}$ and its features $\boldsymbol{\ell}_{i} \in \mathbb{R}^{N^{\prime}_i \times D^{\prime}}$ of all the objects at the latent space are updated as
\begin{equation}
    \setlength\arraycolsep{2pt}
    \begin{bmatrix}
     \boldsymbol{\ell}_{1} & \cdots & \boldsymbol{\ell}_{L}
    \end{bmatrix}
    \leftarrow \mathcal{M}\left(\begin{bmatrix}
        \left(\mathbf{r}_{1}, \boldsymbol{\ell}_{1}\right) & 
        \cdots &
        \left(\mathbf{r}_{L}, \boldsymbol{\ell}_{L}\right) & 
    \end{bmatrix}\right),
\end{equation}
where $L$ represents the total number of objects.

\paragraph{3D occlusion fields.}
Our key insight is that the multi-object encoder primarily learns to avoid collisions between objects and grasp poses in a cluttered scene, as collision modeling requires only local context, making it easier to handle. 
In contrast, occlusion modeling requires a comprehensive understanding of the global context to accurately capture visibility relationships, since occluders and occludees can be positioned far apart. To mitigate this issue, we design 3D occlusion fields that localizes visibility information to voxels via simplified octree-based volume rendering. Concretely, we subdivide a voxel at the latent space into $B^3$ smaller blocks ($B$ blocks per axis), which are projected into the image space. As shown in \Cref{fig:3d_occlusion_fields}, if a block lies within the instance mask corresponding to the target object, a self-occlusion flag $o_{\text{self}}$ is set to $1$. If the block lies within the instance mask of neighbor objects, inter-object occlusion flag $o_{\text{inter}}$ is set to $1$. After computing the flags of all the blocks, we construct the 3D occlusion fields $\mathbfcal{V}_i \in \mathbb{R}^{N^{\prime} \times B^3 \times2}$ by concatenating the two flags of the $i$-th object. 
Finally, we encode it by three layers of 3D CNNs that downsample the resolution by a factor of two at each layer to obtain an occlusion feature $\mathbf{o}_i \in \mathbb{R}^{N^{\prime} \times D^{\prime\prime}}$ at the latent space, and update the latent feature by $
\setlength\arraycolsep{2pt}
\boldsymbol{\ell}_{i} \leftarrow \begin{bmatrix}
\boldsymbol{\ell}_{i} & \mathbf{o}_i
\end{bmatrix}$
to account for occlusions as well as collisions.

\paragraph{Training.} Similar to the standard VAEs~\cite{rempe2021humor,Kingma2013AutoEncodingVB}, we train our model by maximizing the evidence lower bound (ELBO). Additionally, we opt for economic supervision~\cite{wu2025economic} to learn grasp pose prediction efficiently. Therefore, the loss function is defined as
\begin{equation}
    \vspace{-0.1cm}
    \mathcal{L}_{\text{rec}} = \omega_{\text{occ}}\sum^{H}_h \mathcal{L}^h_{\text{occ}} + \omega_{\text{nrm}} \mathcal{L}_{\text{nrm}} + \omega_{\text{SDF}} \mathcal{L}_{\text{SDF}},
    \vspace{-0.2cm}
\end{equation}
\begin{equation}
    \mathcal{L}_{\text{grasp}} = \omega_\text{s}\mathcal{L}_{\text{s}} + \omega_\text{q}\mathcal{L}_{\text{q}} + \omega_a\mathcal{L}_{\text{a}} + \omega_{\text{w}}\mathcal{L}_{\text{w}} + \omega_{\text{d}}\mathcal{L}_{\text{d}},
\end{equation}
\begin{equation}
    \mathcal{L}_{\text{KL}} = \omega_{\text{KL}} D_{\text{KL}}\left(\mathbfcal{E}\left(\mathbf{z}_i \mid \mathbf{x}_i, \mathbf{y}_i\right) \Vert \mathbfcal{P}\left({\boldsymbol\ell}_i, \mathbf{z}_i \mid \mathbf{x}_i\right) \right),
\end{equation}
\begin{equation}
\mathcal{L} = \mathcal{L}_{\text{rec}} + \mathcal{L}_{\text{grasp}} + \mathcal{L}_{\text{KL}},
\end{equation}
where $\mathcal{L}^h_{\text{occ}}$ computes the mean of the binary cross entropy (BCE) function of occupancy at each depth $h$, and $\mathcal{L}_{\text{nrm}}$, and $\mathcal{L}_{\text{SDF}}$ represent the averaged L1 distances of surface normal and SDF, respectively, at the final depth of the octree. $\mathcal{L}_{\text{s}}$, $\mathcal{L}_{\text{q}}$, $\mathcal{L}_{\text{a}}$, $\mathcal{L}_{\text{w}}$, and $\mathcal{L}_{\text{d}}$ computes the averaged L1 distance of graspness of all the possible views, and cross entropy for quality, angle, width, and depth respectively.
Finally, the term $\mathcal{L}_{\text{KL}}$ measures the KL divergence between the encoder $\mathbfcal{E}$ and the prior $\mathbfcal{P}$. Each $\omega$ term is a weight parameter to align the scale of different loss terms.

\begin{figure*}[t]
    \centering
    \scalebox{0.9}{
    \includegraphics[width=\linewidth]{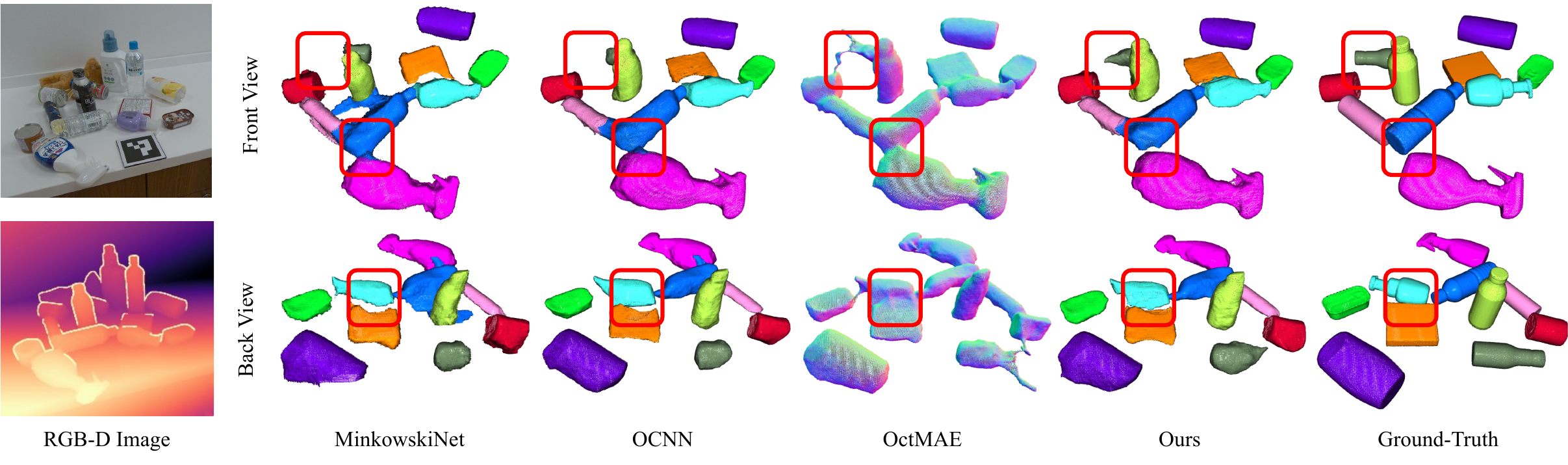}}
    \vspace{-0.4cm}
    \caption{Comparisons of 3D reconstruction methods using sparse voxel representations on the ReOcS dataset. Except OctMAE~\cite{Iwase_ECCV_2024}, an RGB-D image and predicted instance mask are given as input, and the methods output per-object reconstructions. For OctMAE, we visualize its results with normal maps since it is designed to predict a scene-level reconstruction. For a fair comparison, all the models are trained only on the ZeroGrasp-11B dataset. The red rectangles highlight the regions with major differences.}
    \label{fig:results_reconstruction}
\end{figure*}

\begin{table*}[t]
    \caption{Quantitative evaluation of 3D reconstruction on the GraspNet-1B~\cite{fang2020graspnet}, and ReOcS datasets with different difficulties. Chamfer distance (CD), F1-Score@10mm (F1), and normal consistency (NC) are reported in the unit of mm. Seg. denotes an output 3D reconstuction is segmented or not.}
    \centering
    \vspace{-0.3cm}
    \scalebox{0.8}{\begin{tabular}{l|c|ccc|ccc|ccc|ccc}
        \multirow{3}{*}{Method} & \multirow{3}{*}{Seg.} & \multicolumn{3}{c|}{\multirow{2}{*}{GraspNet-1B~\cite{fang2020graspnet}}} & \multicolumn{9}{c}{ReOcS (Ours)} \\ \cline{6-14} \addlinespace[1pt]
        & & & & & \multicolumn{3}{c|}{Easy} & \multicolumn{3}{c|}{Normal} &\multicolumn{3}{c}{Hard} \\ \cline{3-14} \addlinespace[1pt]
        & & CD$\downarrow$ & F1$\uparrow$ & NC$\uparrow$ & CD$\downarrow$ & F1$\uparrow$ & NC$\uparrow$ & CD$\downarrow$ & F1$\uparrow$ & NC$\uparrow$ & CD$\downarrow$ & F1$\uparrow$ & NC$\uparrow$  \\ \hline \addlinespace[1pt]
        Minkowski~\cite{choy20194d} & \checkmark & 6.84 & 81.45 & 77.89 & 5.59 & 85.40 & 84.74 & 6.05 & 82.15 & 82.68 & 9.11 & 77.10 & 80.86 \\
        OCNN~\cite{PSWang2020} & \checkmark & 7.23 & \underline{82.22} & \underline{78.44} & \underline{5.26} & 85.43 & 85.66 & 5.96 & 82.33 & \underline{84.25} & 8.69 & 77.58 & \underline{82.08} \\
        OctMAE~\cite{Iwase_ECCV_2024} & & \underline{7.57} & 78.38 & 75.19 & 5.53 & \underline{87.62} & \textbf{86.90} & \underline{5.93} & \underline{83.98} & 83.45 & \underline{6.76} & \underline{80.24} & 80.58 \\\hline \addlinespace[2pt]
        Ours & \checkmark & \textbf{6.05} & \textbf{84.08} & \textbf{78.46} & \textbf{4.76} & \textbf{88.71} & \underline{86.74} & \textbf{5.54} & \textbf{84.67} & \textbf{85.13} & \textbf{6.73} & \textbf{80.86} & \textbf{82.95} \\
    \end{tabular}}
    \label{tab:main_result_reconstruction}
    \vspace{-0.3cm}
\end{table*}

\subsection{Grasp Pose Refinement}
We find that a strong advantage of 3D reconstruction is its ability to utilize the reconstruction to refine predicted grasp poses. While Ma \etal~\cite{Ma_2024_cvpr} propose a contact-based optimization algorithm, it requires an accurate truncated signed distance field (TSDF) reconstructed from multi-view images and its runtime is relatively slow. In contrast, we introduce a simple refinement algorithm that applies contact-based constraints and collision detection on the 3D reconstruction. Specifically, we first detect contact points by finding the closest points on the reconstruction to the left and right fingers of the gripper. We then adjust the predicted width and depth so that both fingertips have contact. Finally, we perform collision detection with the reconstruction to discard grasp poses with collisions. In the following, we explain the details of these two refinement processes. %

\paragraph{Contact-based constraints.}
Accurate contacts are essential for successful grasping, as they ensure stability and control during manipulation. While our network predicts width and depth of the gripper, we observe that even small errors can result in unstable grasping. To address this issue, we refine a grasp pose by adjusting the fingertip locations of the gripper to align with the nearest contact points of the left and right fingers $\mathbf{c_{\text{L}}}$ and $\mathbf{c_{\text{R}}}$ on the reconstruction. Based on the contact points the width $\mathbf{w}$ is refined as
\begin{equation}
     \Delta \mathbf{w}  = \min \left( D\left(\mathbf{c_{\text{L}}}\right),  D\left(\mathbf{c_{\text{R}}}\right)\right),
\end{equation}
\begin{equation}
\label{eq:update_width}
\mathbf{w} \leftarrow \mathbf{w} + 2 \left( \max\left(\gamma_{\text{min}}, \min(\Delta \mathbf{w}, \gamma_{\text{max}})\right) - \Delta \mathbf{w} \right),
\end{equation}
so that the contact distance $\Delta \mathbf{w}$ remains within the range $\gamma_{\text{min}}$ to $\gamma_{\text{max}}$. %
Note that $D\left(\mathbf{c}\right)$ denotes the contact distance from $\mathbf{c}$.
We further adjust the depth $\mathbf{d}$ by
\begin{equation}
\label{eq:update_depth}
    \mathbf{d} \leftarrow \max\left(Z\left(\mathbf{c}_{\text{L}}\right), Z\left(\mathbf{c}_{\text{R}}\right)\right),
\end{equation}
where $Z\left(\mathbf{c}\right)$ compute depth of the contact point $\mathbf{c}$. These simple refinement steps help ensure stable grasps.

\paragraph{Collision detection.}
We implement a simple model-free collision detector using the two-finger gripper, following GS-Net~\cite{Wang_2021_ICCV}. Although the previous method uses partial point cloud obtained from a depth map, it fails to discard predicted grasp poses that result in collisions with occluded regions. To overcome this limitation, we instead leverage the reconstructed shapes, which allows more precise collision detection. To justify this approach, we perform extensive analysis in our experiments and show the advantages.

\begin{figure*}[t]
    \centering
    \includegraphics[width=0.9\linewidth]{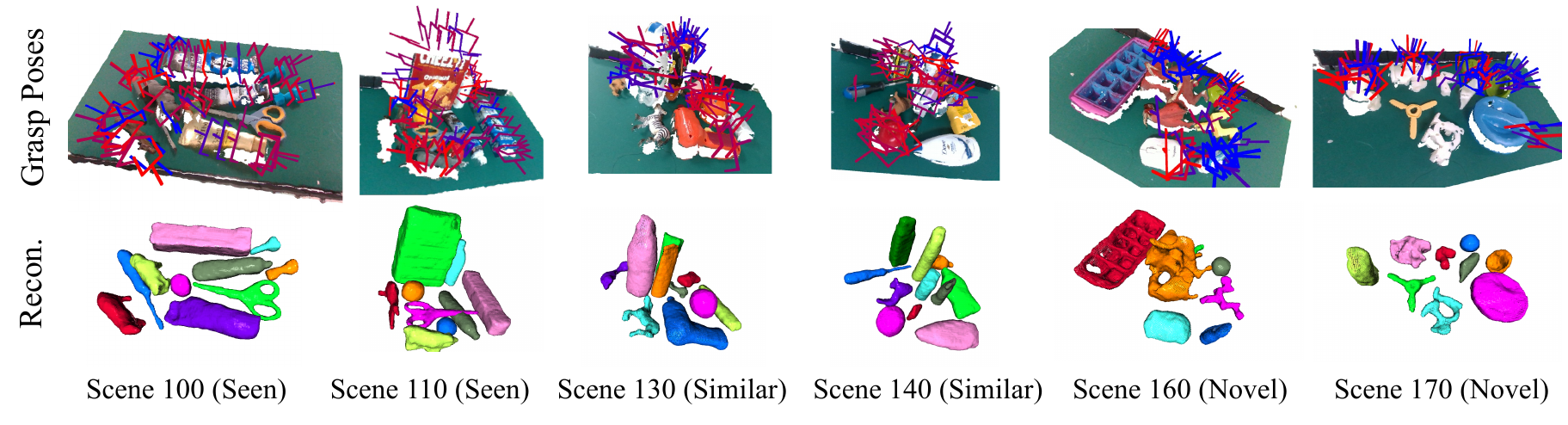}
    \caption{Qualitative results on grasp pose prediction of ZeroGrasp. Following GSNet~\cite{Wang_2021_ICCV}, we show the best 50 grasp predictions after grasp-NMS~\cite{fang2020graspnet} from six different scenes (two scenes per split). Red and blue grasps denote high and low grasp quality scores, respectively.}
    \label{fig:results_grasp}
\end{figure*}

\begin{table*}[t]
    \begin{minipage}{0.6\linewidth}
    \caption{Quantitative evaluation of grasp pose prediction on the GraspNet-1Billion benchmark. Similar to the other baseline methods, we report the average precision (AP), AP$_{0.4}$, and AP$_{0.8}$. Note that $0.4$ and $0.8$ denote the friction coefficients, and the lower the more difficult. G and R in the output column indicate whether the reconstructed posture prediction are predicted or not.}
    \centering
    \scalebox{0.6}{\begin{tabular}{l|cc|ccc|ccc|ccc}
        \multirow{2}{*}{Method} & \multicolumn{2}{c|}{Output} & \multicolumn{3}{c|}{Seen} & \multicolumn{3}{c|}{Similar} &\multicolumn{3}{c}{Novel} \\ \cline{2-12} \addlinespace[1pt]
        & G & R & \textbf{AP} & AP$_{0.8}$ & AP$_{0.4}$ & \textbf{AP} & AP$_{0.8}$ & AP$_{0.4}$ & \textbf{AP} & AP$_{0.8}$ & AP$_{0.4}$ \\ \hline \addlinespace[1pt]
        GG-CNN~\cite{morrison2018closing} & \checkmark & & $15.48$ & $21.84$ & $10.25$ & $13.26$ & $18.37$ & $4.62$ & $5.52$ & $5.93$ & $1.86$ \\
        Chu \etal~\cite{chu2018deep} & \checkmark & & $15.97$ & $23.66$ & $10.80$ & $15.41$ & $20.21$ & $7.06$ & $7.64$ & $8.69$ & $2.52$ \\
        CenterGrasp$^{\dagger}$~\cite{chisari2024centergrasp} & \checkmark & \checkmark & $16.46$ & $20.24$ & $11.74$ & $9.52$ & $11.92$ & $5.71$ & $1.60$ & $1.89$ & $1.12$ \\
        GPD~\cite{Pas2017GraspPD} & \checkmark & & $22.87$ & $28.53$ & $12.84$ & $21.33$ & $27.83$ & $9.64$ & $8.24$ & $8.89$ & $2.67$ \\
        Lian \etal~\cite{liang2019pointnetgpd} & \checkmark & & $25.96$ & $33.01$ & $15.37$ & $22.68$ & $29.15$ & $10.76$ & $9.23$ & $9.89$ & $2.74$ \\
        GraspNet~\cite{fang2020graspnet} & \checkmark & & $27.56$ & $33.43$ & $16.59$ & $26.11$ & $34.18$ & $14.23$ & $10.55$ & $11.25$ & $3.98$ \\
        GSNet~\cite{Wang_2021_ICCV} & \checkmark & & $67.12$ & $78.46$ & $60.90$ & $54.81$ & $ 66.72$ & $46.17$ & $24.31$ & $30.52$ & $14.23$ \\
        Ma \etal~\cite{Ma_2022_CoRL} & \checkmark & & $63.83$ & $74.25$ & $58.66$ & $58.46$ & $70.05$ & $51.32$ & $24.63$ & $31.05$ & $12.85$ \\
        HGGD & \checkmark & & 64.45 & 72.81 & 61.16 & 53.59 & 64.12 & 45.91 & 24.59 & 30.46 & \underline{15.58} \\
        EconomicGrasp~\cite{wu2025economic} & \checkmark & & 68.21 & 79.60 & 63.54 & 61.19 & 73.60 & 53.77 & 25.48 & 31.46 & 13.85 \\ \hline \addlinespace[2pt]
        Ours & \multirow{2}{*}{\checkmark} & \multirow{2}{*}{\checkmark} & $\underline{70.53}$ & $\underline{82.28}$ & $\underline{64.26}$ & $\underline{62.51}$ & $\underline{74.26}$ & $\underline{54.97}$ & $\underline{26.46}$ & $\underline{33.13}$ & $15.11$ \\
        Ours+FT & & & $\mathbf{72.43}$ & $\mathbf{83.12}$ & $\mathbf{65.57}$ & $\mathbf{65.45}$ & $\mathbf{78.32}$ & $\mathbf{55.48}$ & $\mathbf{28.49}$ & $\mathbf{34.21}$ & $\mathbf{15.80}$ \\
   \end{tabular}}
    \label{tab:main_result_grasp}
    \end{minipage} \hspace{0.2cm}
    \begin{minipage}{0.35\linewidth}
        \caption{Abalations on the network input, architecture, and refinement algorithm. For reconstruction and grasp pose prediction, we report the metrics of the hard split of the ReOcS dataset and GraspNet-1B dataset, respectively.}
    \centering
    \scalebox{0.58}{\begin{tabular}{l|ccc|ccc}
        \multirow{2}{*}{Method} & \multicolumn{3}{c|}{Reconstruction} & \multicolumn{3}{c}{Grasp Pose} \\ \cline{2-7} \addlinespace[1pt]
        & CD$\downarrow$ & F1$\uparrow$ & NC$\uparrow$ & Seen & Similar & Novel \\ \hline \addlinespace[1pt]
        Baseline (OCNN~\cite{10.1145/3072959.3073608}) & 8.69 & 77.58 & 82.08 & 41.27 & 36.48 & 17.46 \\ \hline \addlinespace[1pt]
        No CVAE & 7.67 & 78.79 & 82.35 & 70.23 & 60.31 & 26.28 \\ \addlinespace[1pt]
        No Multi-Obj. Encoder & 7.09 & 79.62 & 82.60 & 69.52 & 61.03 & 26.17 \\ \addlinespace[1pt]
        No 3D Occlusion Fields & 7.54 & 78.81 & 81.94 & 67.34 & 58.45 & 25.00 \\ \addlinespace[1pt]
        No Contact Constraints & 6.73 & 80.86 & 82.95 & 65.67 & 55.34 & 24.92 \\ \addlinespace[1pt]
        No Collision Detection & 6.73 & 80.86 & 82.95 & 49.35 & 44.28 & 21.03 \\ \addlinespace[1pt]
        + Depth Map & 6.73 & 80.86 & 82.95 & 59.93 & 51.58  & 24.07 \\ \hline \addlinespace[1pt]
        Ours & 6.73 & 80.86 & 82.95 & 70.53 & 62.51 & 26.46 \\ \addlinespace[1pt]
    \end{tabular}}
    \label{tab:ablation}
    \end{minipage}
\end{table*}

\section{Datasets}
We create two datasets for evaluation and training --- 1) the ReOcS dataset is designed to evaluate the quality of 3D reconstruction under varying occlusion levels, and 2) the ZeroGrasp-11B dataset is intended for training baselines and our model for zero-shot robotic grasping. \Cref{fig:dataset} highlights several examples of the datasets.

\subsection{ReOcS Dataset}
The ReOcS dataset contains 1,125 RGB-D images and ground-truth instance masks, 6D poses, and 3D shapes. To obtain accurate depth maps of metalic and transparent objects, we use a learning-based stereo depth estimation algorithm~\cite{Shankar2021ALS}. There are three splits --- easy, normal and hard --- based on the extent of occlusions. We use this dataset to compare the robustness of baselines and our method under different occlusion conditions. For the details, please refer to the supplementary material.

\subsection{ZeroGrasp-11B Dataset}
As shown in \Cref{tab:dataset}, the ZeroGrasp-11B dataset leverages $12$K 3D models and create $1$M photorealistic RGB images, ground-truth and stereo depth maps of 25,000 scenes with BlenderProc~\cite{Denninger2023}. In addition, it provides ground-truth 3D reconstructions and 6D object poses. While Grasp-Anything-6D~\cite{nguyen2024language} has 6D annotations of a larger number of objects, 3D models are missing, which is crucial for reconstruction. Further, its synthesized images and predicted depth maps have no guarantee that they are physically valid, and grasp pose annotations are sparse and generated from planar grasp poses. We solve these issues with the ZeroGrasp-11B dataset to enable zero-shot robotic grasping. In the following, we describe the procedure of grasp pose generation.

\paragraph{Grasp pose generation.}
Following \cite{mousavian2019graspnet}, we begin by randomly sampling $N_s$ surface points on ground-truth 3D reconstructions. $N_s$ is determined by $N_s = \mathcal{A} / \rho$ with $\mathcal{A}$ denoting the surface area and $\rho$ as a density parameter. For each surface point, we synthesize candidate grasps with all combinations of views, orientations around the point's normal vector, and depth respectively, following GraspNet-1B~\cite{fang2020graspnet}. Next, we conduct collision detection to eliminate any grasps with collision and compute the grasp quality $\mathbf{q}$ for the remaining candidates. The quality metric~\cite{1087483} is computed based on the normal vectors $\mathbf{n}_{L}$ and $\mathbf{n}_{R}$ of the contact points $\mathbf{c}_{L}$ and $\mathbf{c}_{R}$ by $\mathbf{q} = \min \left(\mathbf{n}_{L} \cdot \mathbf{c}_{LR}, \, \mathbf{n}_{R} \cdot \mathbf{c}_{LR}  \right),$
where $\mathbf{c}_{LR} = \left(\mathbf{c}_{L} - \mathbf{c}_{R}\right)/\left(\Vert\mathbf{c}_{L}\Vert\Vert\mathbf{c}_{R}\Vert\right)$.
Finally, we physically validate the generated grasps with IsaacGym~\cite{DBLP:journals/corr/abs-2108-10470}. To make the Objaverse 3D models compatible with simulation, we decompose them into convex hulls using V-HACD~\cite{mamou_vhacd}. \Cref{fig:dataset} shows the grasp poses before and after the collision and physics-based filtering process.

\section{Experiments}

\paragraph{Implementation details.}
Our propoposed method, ZeroGrasp, adopts a ResNeXt~\cite{Xie2016} architecture, pretrained on the ImageNet dataset~\cite{deng2009imagenet}, as an image encoder, and all the parameters except the last layer are fixed during training. Similar to EconomicGrasp~\cite{wu2025economic}, we use the predicted view graspness $\mathbf{s}$ to determine a view direction. For training, we use AdamW~\cite{loshchilov2018decoupled} with a learning rate of $0.001$, batch size of $16$ on NVIDIA A100. The weights of the loss function is provided in the supplemental. We set the dimensions of the input image feature $D$, the latent feature $D^{\prime}$, and the 3D occlusion fields $\mathcal{V}$ to $32$, $192$, and $16$ respectively. For the 3D occlusion fields, we use $8$ for the block resolution $B$. Following Ma \etal, the ranges of the contact distance $\gamma_{\text{min}}$ and $\gamma_{\text{max}}$ are defined to $0.005$m and $0.02$m, respectively. To generate grasp poses, we use $0.005\text{m}^2$ as the sampling density $\rho$.

\paragraph{Metrics.}
Similar to OctMAE~\cite{Iwase_ECCV_2024}, we use the Chamfer distance (CD), F-1 score, and normal consistency (NC) to evaluate the quality of 3D reconstruction. To evaluate the quality of grasp pose prediction, we use average precision (AP), a standard metric of the GraspNet-1B benchmark, which evaluates average precision based on the top-k ranked grasps in a scene. The AP$_\mu$ metric measures the precision with the friction of $\mu$ by evaluating grasps with friction $\mu$ over m different thresholds. The final AP score is computed as the mean of AP$_\mu$, using friction values $\mu$ from $0.2$ to $1.2$ at intervals of $0.2$.

\subsection{Main Results}

\paragraph{3D reconstruction.}
 As shown in \Cref{tab:main_result_reconstruction}, our method outperforms the other single-view reconstruction methods. We choose the three methods using sparse voxel representations due to its superior efficiency and accuracy in a zero-shot setup, reported in Iwase \etal~\cite{Iwase_ECCV_2024}. We train the baseline and our methods on the ZeroGrasp-11B dataset and evaluate them on the GraspNet-1B and ReOcS dataset to test generalization to real-world images. Our qualitative evaluation in \Cref{fig:results_reconstruction} demonstrates the robustness of ZeroGrasp to real-world images and inter-object occlusions. 

\vspace{-0.3cm}
\paragraph{Grasp pose prediction.}
\Cref{tab:main_result_grasp} demonstrates the comparison against state-of-the-art methods for grasp pose prediction on the RealSense data of the GraspNet-1Billion benchmark. The baselines and our model are trained on the training split of the GraspNet-1Billion dataset for 20 epochs. Notably, our method achieves the state-of-the-art performance across all the AP metrics.
In the Ours+FT setup, our model is initially pre-trained on the ZeroGrasp-11B dataset, then fine-tuned on the GraspNet-1Billion dataset for 2 epochs. As a result, fine-tuning improves $1.9$\%, $2.94$\%, and $2.03$\% in the seen, similar and novel splits. This result supports the importance of large-scale grasp pose datasets for zero-shot robotic grasping. \Cref{fig:results_grasp} shows qualitative results of ZeroGrasp. Unlike the previous methods, ZeroGrasp enables accurate grasp pose prediction even in occluded or truncated regions.

\subsection{Ablations}
\Cref{tab:ablation} shows our ablation studies to validate the effectiveness of each component. We provide detailed analyses from the perspectives of the two tasks addressed in our work.

\vspace{-0.3cm}
\paragraph{3D reconstruction.}
We observe a consistent drop in performance across all reconstruction metrics when each of CVAE, the multi-object encoder, and 3D occlusion fields is individually excluded. This highlights the importance of multi-object reasoning to achieve higher reconstruction quality. As shown in \Cref{fig:results_reconstruction}, our visualizations further demonstrate that these components contribute to better reconstruction, especially in regions with inter-object occlusions and close contacts between objects.

\vspace{-0.3cm}
\paragraph{Grasp pose prediction.}
As illustrated in \Cref{tab:ablation}, most of the components contribute to improved grasp pose detection. In particular, collision detection and contact-based constraints provide a significant boost to grasp pose quality. Our comparison of collision detection using a depth map (partial point clouds) as in GSNet~\cite{Wang_2021_ICCV} and our predicted reconstruction ($59.93$ vs $70.53$) reveals that reconstruction-based collision detection is more effective. Furthermore, the substantial performance drop without 3D occlusion fields underscores the importance of reasoning about inter-object occlusions.

\begin{figure}
    \centering
    \includegraphics[width=0.95\linewidth]{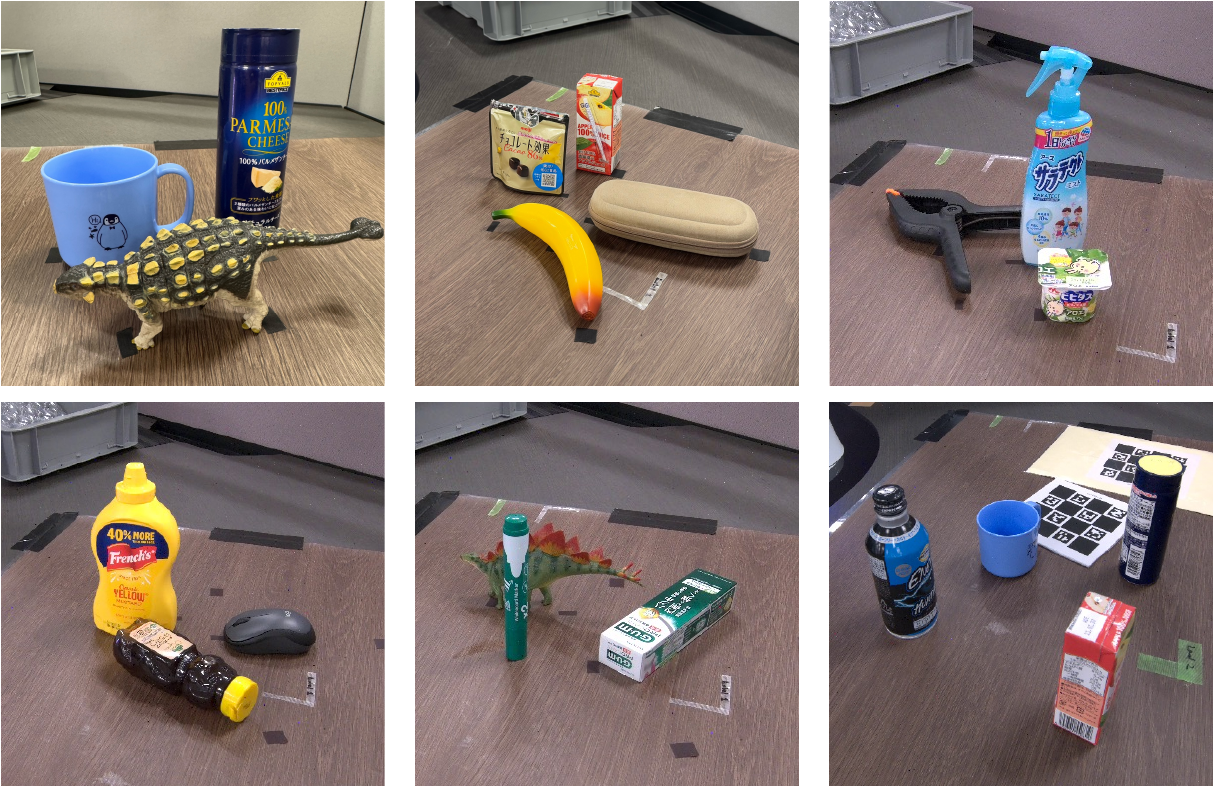}
    \caption{Example scenes of our real-robot evaluation.}
    \label{fig:real-robot}
    \vspace{-0.6cm}
\end{figure}

\vspace{-0.2cm}
\subsection{Real-Robot Evaluation}
We validate the feasibility and generalizability of the baseline (OCNN~\cite{10.1145/3072959.3073608}) and our method, trained only on our synthetic dataset, through real-world evaluations. Our robotic setup uses Franka Emika Panda robot and Robotiq 2F-85 hand. As shown in \Cref{fig:real-robot}, we set up $5$ scenes with $3$ to $4$ objects. Each object is picked up in repeated trials, with a maximum of $3$ attempts per object. Our success rate, measured by the ratio of objects successfully picked up, is $56.25$\% for the baseline and $75$\% for our method, highlighting the strong generalization of our approach in real-world scenarios. We describe more details about the robotic setup and show qualitative results in the supplementary material.

\vspace{-0.2cm}
\section{Conclusion}
In this paper, we propose ZeroGrasp, a novel approach for simultaneous 3D reconstruction and grasp pose prediction. By integrating five key components, ZeroGrasp enhances both shape reconstruction and grasp prediction quality. Our extensive analysis confirms the effectiveness of these components.
In addition, we strongly believe that ZeroGrasp-11B dataset facilitates future research in zero-shot robotic grasping.
Despite its promising generalization capabilities, ZeroGrasp has some limitations. First, our method does not support incremental or multi-view 3D reconstruction~\cite{murORB2,mildenhall2020nerf}, which is beneficial when using a wrist-mounted camera on an end effector. Second, it does not account for placement poses that could leverage predicted 3D reconstructions. While this paper focuses on single-view 3D reconstruction and grasp pose prediction, exploring these directions would be valuable.

\vspace{-0.2cm}
\section*{Acknowledgment}
We thank Tianyi Ko for help with the real-world robot experiments. This research was supported by Toyota Research Institute.

\bibliographystyle{ieeetr}
\bibliography{main}

\clearpage
\appendix

\section{Dataset Details}
\begin{table}[h]
    \caption{The number of images and 3D models across the different splits of the ReOcS dataset.}
    \centering
    \setlength{\tabcolsep}{8pt}
    \scalebox{0.8}{
    \begin{tabular}{l|cc}
       Split & \# Images & \# 3D Models \\ \hline
       Easy & 479 & 22 \\
       Normal & 329 & 22 \\
       Hard & 317 & 20 \\ \hline
       Total & 1125 & 22 \\ \hline
    \end{tabular}}
    \label{tab:reocs_dataset}
\end{table}

\begin{figure*}[t]
    \centering
    \includegraphics[width=1.0\linewidth]{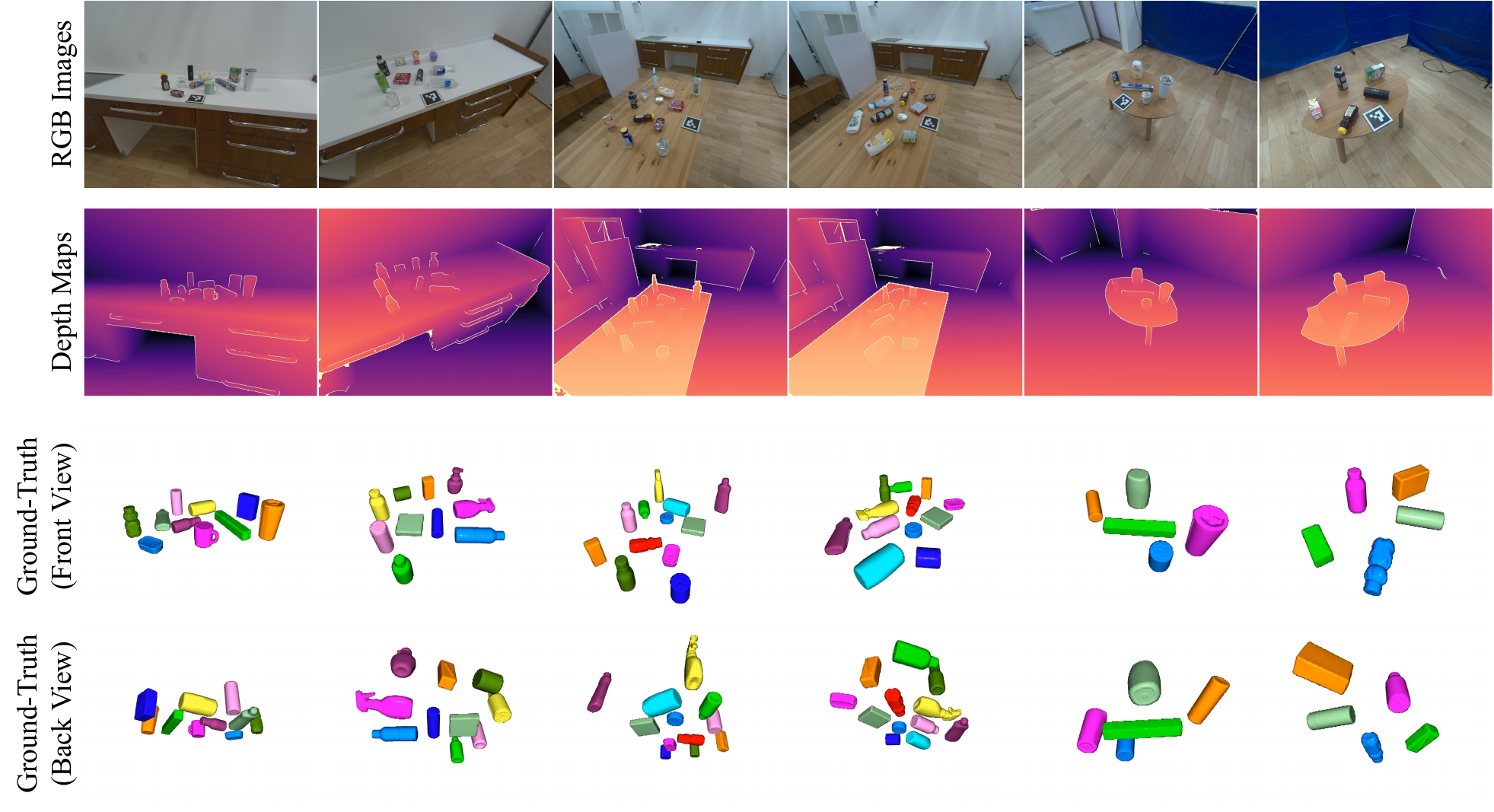}
    \caption{Examples from the easy split of the ReOcS dataset, where ground-truth 3D models are represented as octrees derived from their corresponding meshes.}
    \label{fig:reocs_easy}
\end{figure*}

\begin{figure*}[t]
    \centering
    \includegraphics[width=1.0\linewidth]{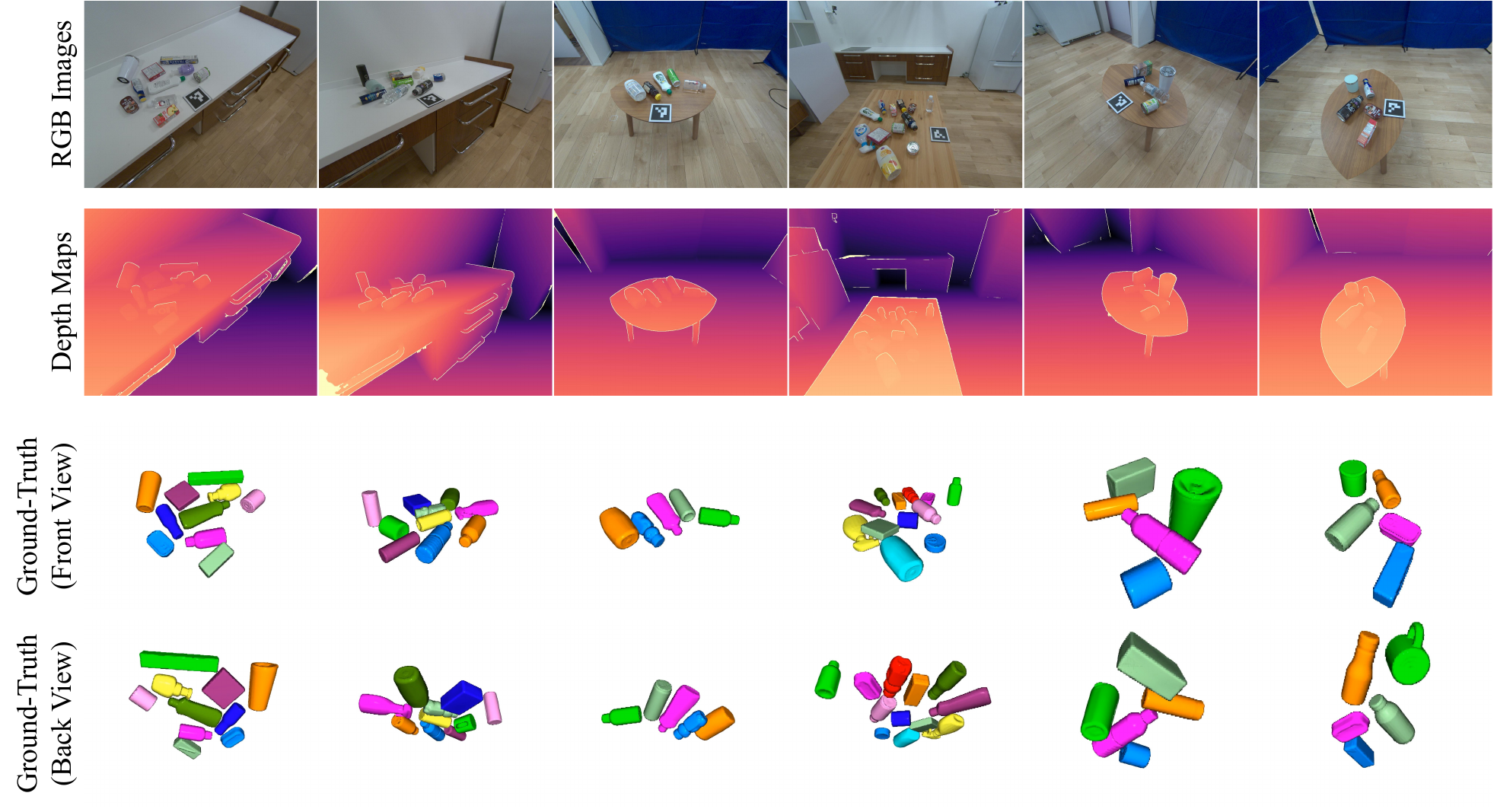}
    \caption{Examples from the normal split of the ReOcS dataset, where ground-truth 3D models are represented as octrees derived from their corresponding meshes.}
    \label{fig:reocs_normal}
\end{figure*}

\begin{figure*}[t]
    \centering
    \includegraphics[width=1.0\linewidth]{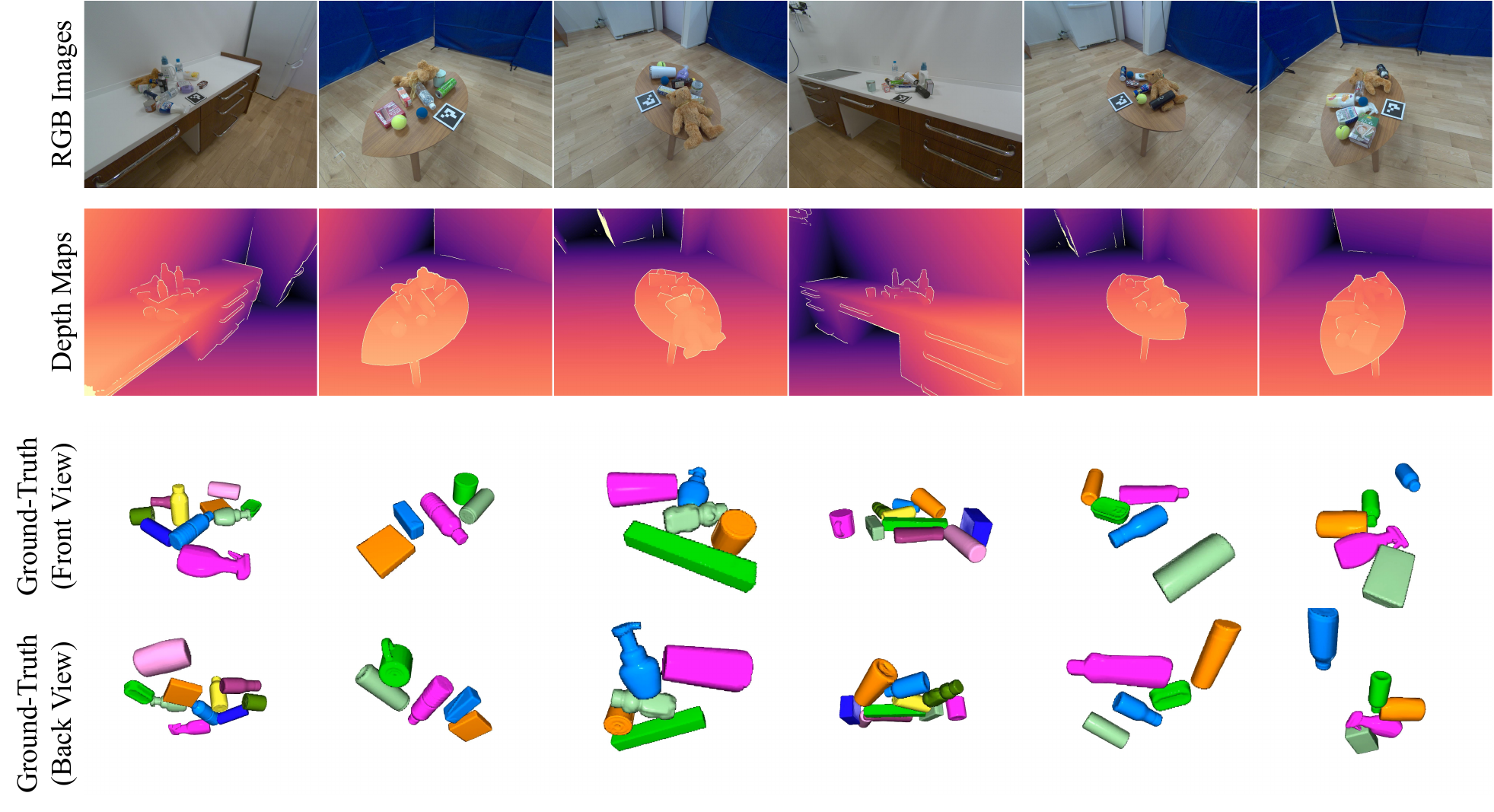}
    \caption{Examples from the hard split of the ReOcS dataset, where ground-truth 3D models are represented as octrees derived from their corresponding meshes.}
    \label{fig:reocs_hard}
\end{figure*}

\begin{figure*}[t]
    \centering
    \includegraphics[width=1.0\linewidth]{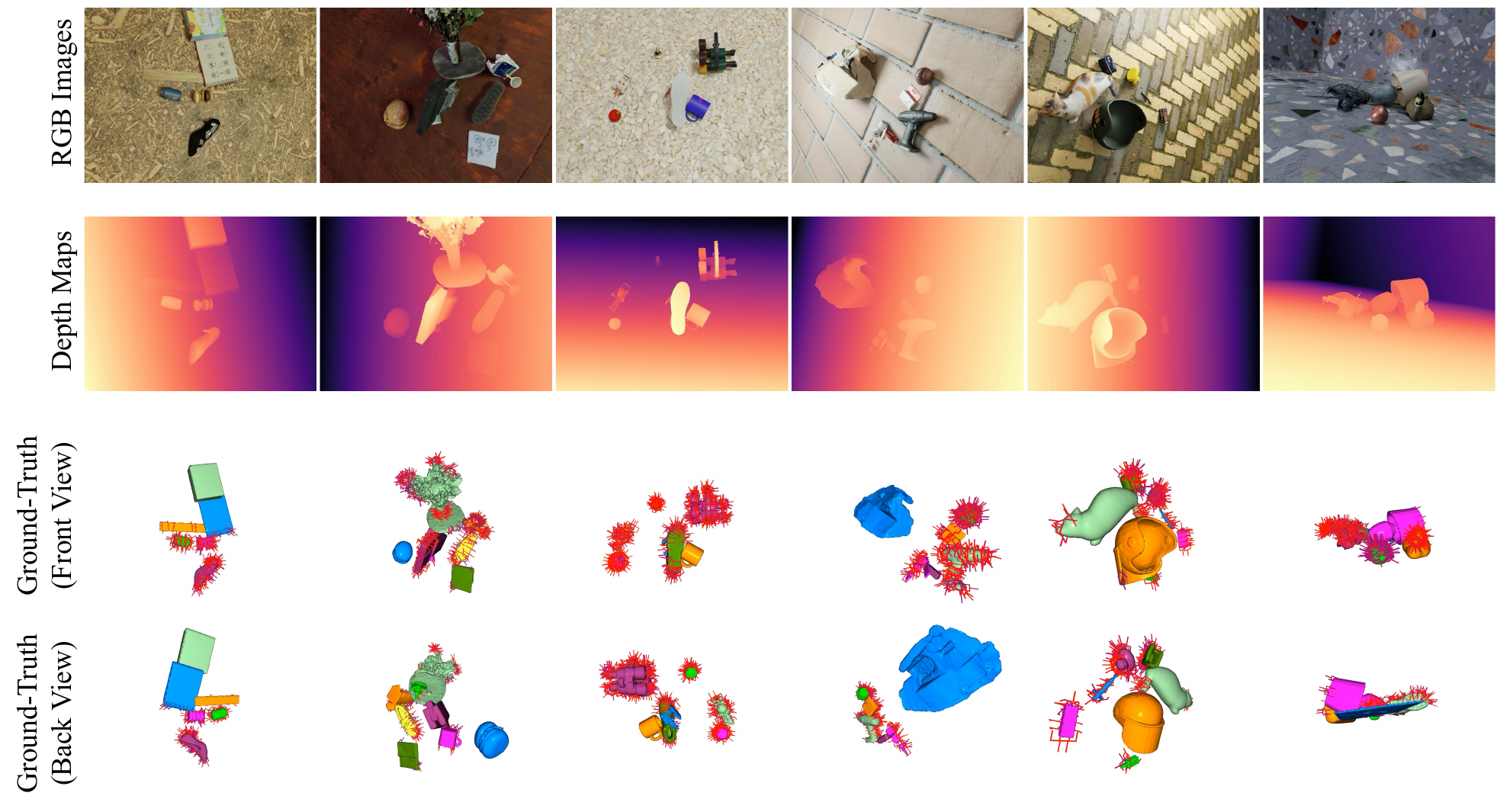}
    \caption{Examples from the ZeroGrasp-11B dataset, where ground-truth 3D models and grasp poses are represented as octrees derived from their corresponding meshes and two-finger parallel grippers.}
    \label{fig:zerograsp_11b_0}
\end{figure*}

\begin{figure*}[t]
    \centering
    \includegraphics[width=1.0\linewidth]{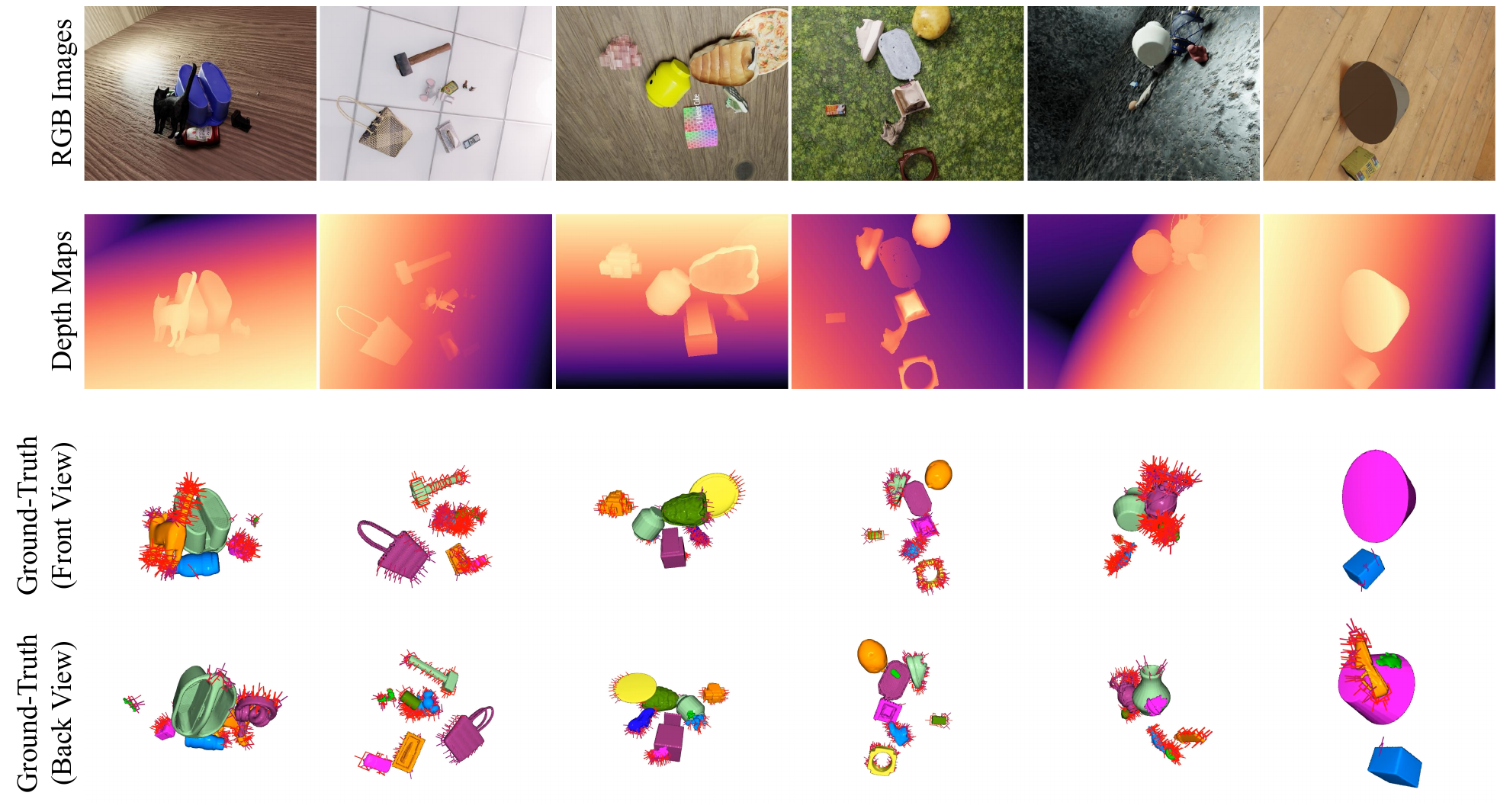}
    \caption{Examples from the ZeroGrasp-11B dataset, where ground-truth 3D models and grasp poses are represented as octrees derived from their corresponding meshes and two-finger parallel grippers.}
    \label{fig:zerograsp_11b_1}
\end{figure*}

\begin{figure*}[t]
    \centering
    \includegraphics[width=1.0\linewidth]{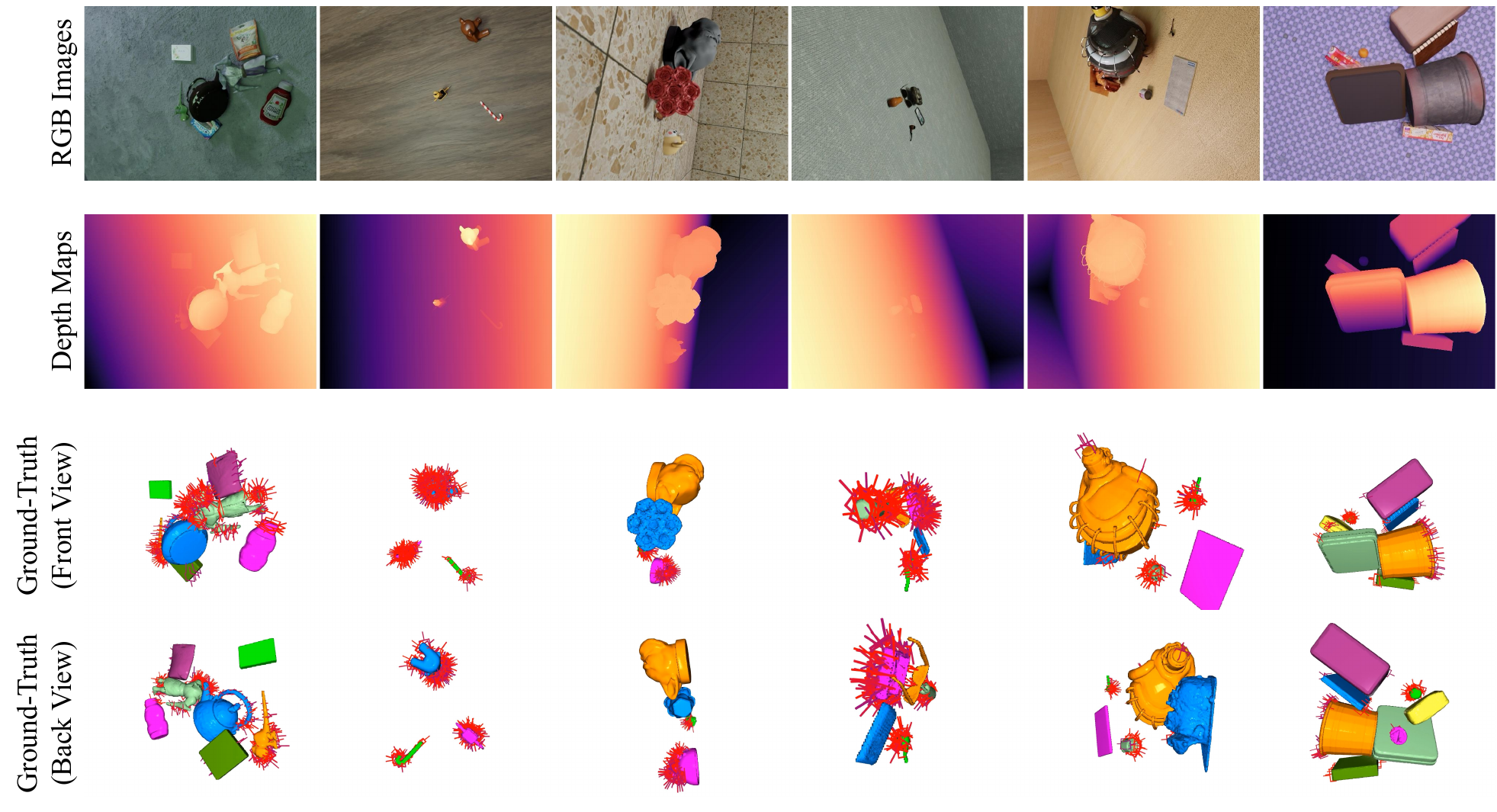}
    \caption{Examples from the ZeroGrasp-11B dataset, where ground-truth 3D models and grasp poses are represented as octrees derived from their corresponding meshes and two-finger parallel grippers.}
    \label{fig:zerograsp_11b_2}
\end{figure*}

\Cref{tab:reocs_dataset} shows the number of images and 3D models of the three splits --- easy, normal and hard --- of the ReOcS dataset. We show examples from the ReOcS and ZeroGrasp-11B datasets in \Cref{fig:reocs_easy,fig:reocs_normal,fig:reocs_hard} and \Cref{fig:zerograsp_11b_0,fig:zerograsp_11b_1,fig:zerograsp_11b_2}, respectively. For physics-based grasp pose validation on the ZeroGrasp-11B dataset, we use the Franka Emika Panda hand. We first move the gripper given a synthesized grasp pose and apply a closing force of $80$N. Next, we swing the gripper orthogonally to the direction of the parallel fingers by $60$ degrees, repeating this motion four times. A grasp is considered successful only if both fingers maintain contact after the motion. For instance masks, we fine-tune SAM-2 on the ZeroGrasp-11B dataset and the training split of the GraspNet-1B dataset.

\section{Experiments}

\subsection{Implementation Details}
We compute the final grasp scores by multiplying the predicted graspness $\mathbf{g}$ and quality $\mathbf{q}$ metrics. The predicted width and depth are clipped to lie within the ranges $[0.0, 0.10]$ and $[0.0, 0.04]$, respectively. The learning rate is decayed by a factor of $0.5$ every $5000$ iterations. The weight parameters of the loss function are set as follows; $\omega_{\text{occ}} = 1.0$, $\omega_{\text{nrm}} = 1.0$, $\omega_{\text{SDF}} = 1.0$, $\omega_{\text{g}} = 5.0$, $\omega_{\text{q}} = 5.0$, $\omega_{\text{a}} = 1.0$, $\omega_{\text{t}} = 1.0$, $\omega_{\text{w}} = 2.5$, $\omega_{\text{d}} = 2.5$, and $\omega_{\text{KL}} = 0.5$. We use three layers of 3D CNNs that downsample the resolution by a factor of two at each layer to extract 1-D features from 3D occlusion fields per voxel in the latent space. We use the training split of GraspNet-1B to fine-tune SAM2-tiny.

\subsection{Metrics}
Similar to OctMAE~\cite{Iwase_ECCV_2024}, we evaluate the reconstruction metrics such as Chamfer distance (CD), F-1 score and normal consistency (NC) with the following equations. Note that the predicted surface points $\mathcal{P}_{\text{pd}}$ are derived from the predicted occupied points, normal vectors, and SDF. The ground-truth surface points is denoted as $\mathcal{P}_{\text{gt}}$. 

\paragraph{Chamfer distance (CD).}
We use the bidirectional Chamfer distance as a more balanced metric to measure the similarity between two point cloud sets.
\begin{equation}
\begin{aligned}
\text{CD}(\mathcal{P}_{\text{pd}}, \mathcal{P}_{\text{gt}}) &= \frac{1}{2|\mathcal{P}_{\text{pd}}|} \sum_{\mathbf{x}_{\text{pd}} \in \mathbf{\mathcal{P}_{\text{pd}}} } \min_{\mathbf{x}_{\text{gt}} \in \mathbf{\mathcal{P}_{\text{gt}}}} ||\mathbf{x}_{\text{pd}} - \mathbf{x}_{\text{gt}}|| \\
&+ \frac{1}{2|\mathcal{P}_{\text{gt}}|} \sum_{\mathbf{x}_{\text{gt}} \in \mathbf{\mathcal{P}_{\text{gt}}} } \min_{\mathbf{x}_{\text{pd}} \in \mathbf{\mathcal{P}_{\text{pd}}}} ||\mathbf{x}_{\text{gt}} - \mathbf{x}_{\text{pd}}||.
\end{aligned}
\end{equation}

\paragraph{F-1 score.}
We use the threshold $\eta$ of $1$ cm for all evaluations, in regardless of the object size.
\begin{equation}
\begin{aligned}
P & =\frac{\left|\left\{\mathbf{x}_{\mathrm{pd}} \in \mathcal{P}_{\mathrm{pd}} \mid \min_{\mathbf{x}_{\mathrm{gt}} \in \mathcal{P}_{ \mathrm{gt}}}\left\|\mathbf{x}_{\mathrm{gt}}-\mathbf{x}_{\mathrm{pd}}\right\|<\eta\right\}\right|}{\left|\mathcal{P}_{\mathrm{pd}}\right|}, \\
R & =\frac{\left|\left\{\mathbf{x}_{\mathrm{gt}} \in \mathcal{P}_{\mathrm{gt}}\mid \min_{\mathbf{x}_{\mathrm{pd}} \in \mathcal{P}_{\mathrm{pd}}} \left\| \mathbf{x}_{\mathrm{pd}}-\mathbf{x}_{\mathrm{gt}} \right\|<\eta\right\}\right|}{\left|\mathcal{P}_{\mathrm{gt}}\right|}, \\
\end{aligned}
\end{equation}

\begin{equation}
    \text{F-1 score} = \frac{2 P R}{P + R}.
\end{equation}

\paragraph{Normal consistency (NC).}
We use the symmetrized normal consistency metric, borrowed from ConvONet~\cite{Peng2020ECCV}. 
\begin{equation}
\begin{aligned}
\text{NC}(\mathbf{N}_{\text{pd}}, \mathbf{N}_{\text{gt}}) = \frac{1}{2|\mathbf{N}_{\text{pd}}|} \sum_{\mathbf{n}_{\text{pd}} \in \mathbf{\mathbf{N}_{\text{pd}}}} \left(\mathbf{n}_{\text{pd}} \cdot \mathbf{n}^{*}_{\text{gt}}\right) \\
+ \frac{1}{2|\mathbf{N}_{\text{gt}}|} \sum_{\mathbf{n}_{\text{gt}} \in \mathbf{\mathbf{N}_{\text{gt}}} } \left(\mathbf{n}_{\text{gt}} \cdot \mathbf{n}^{*}_{\text{pd}}\right),
\end{aligned}
\end{equation}
where  $\mathbf{n}^{*}_{\text{gt}}$ and $\mathbf{n}^{*}_{\text{pd}}$ represent the closest normal vectors, respectively.

\subsection{Qualitative Results}
\begin{figure*}[t]
    \centering
    \includegraphics[width=0.93\linewidth]{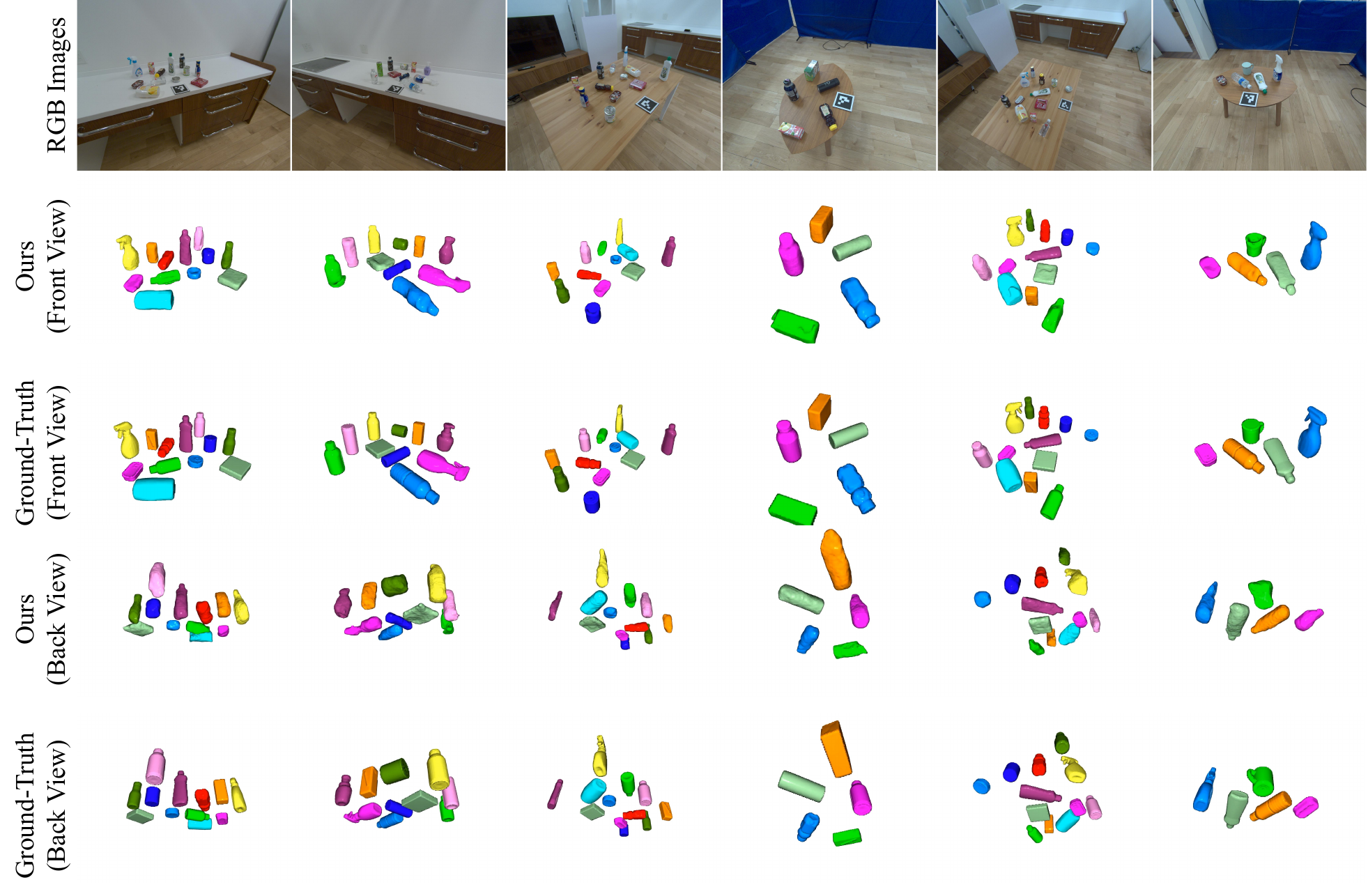}
    \vspace{-0.3cm}
    \caption{Results of 3D reconstruction on the easy split of the ReOcS dataset.}
    \label{fig:recon_reocs_easy}
\end{figure*}

\begin{figure*}[t]
    \centering
    \includegraphics[width=0.93\linewidth]{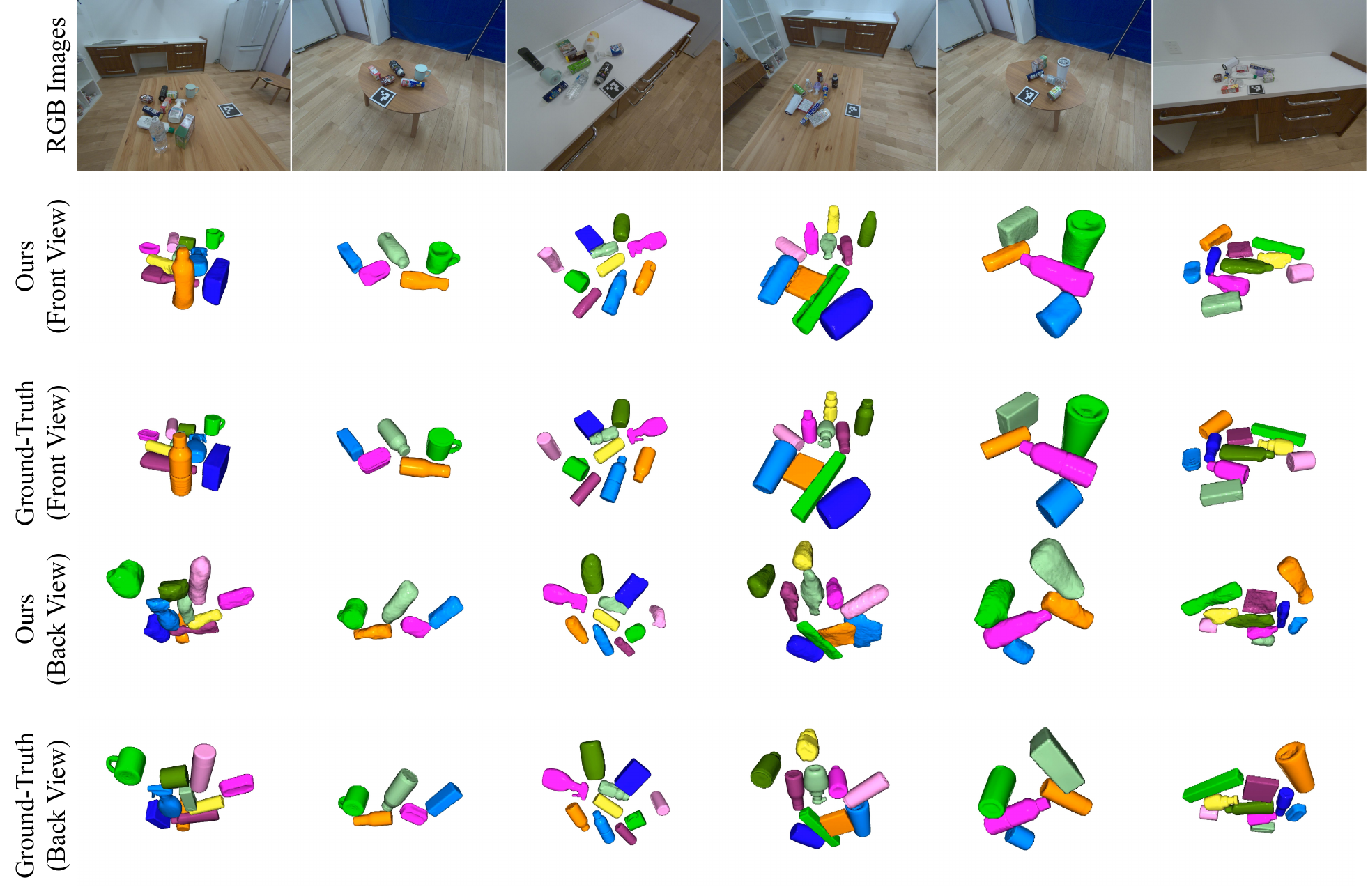}
    \vspace{-0.3cm}
    \caption{Results of 3D reconstruction on the normal split of the ReOcS dataset.}
    \label{fig:recon_reocs_normal}
\end{figure*}

\begin{figure*}[t]
    \centering
    \includegraphics[width=0.93\linewidth]{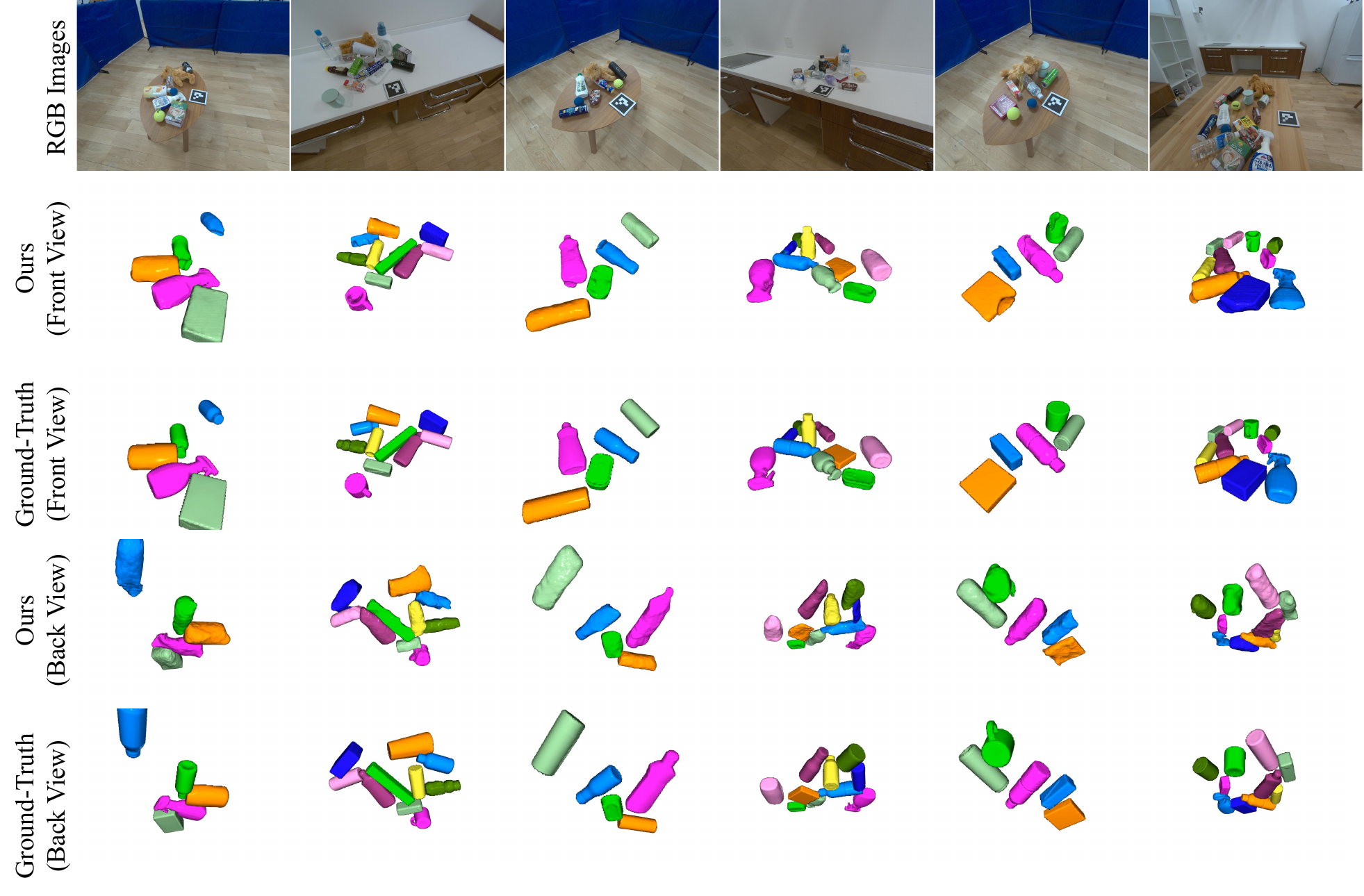}
    \vspace{-0.3cm}
    \caption{Results of 3D reconstruction on the hard split of the ReOcS dataset.}
    \label{fig:recon_reocs_hard}
\end{figure*}

\begin{figure*}[t]
    \centering
    \includegraphics[width=0.81\linewidth]{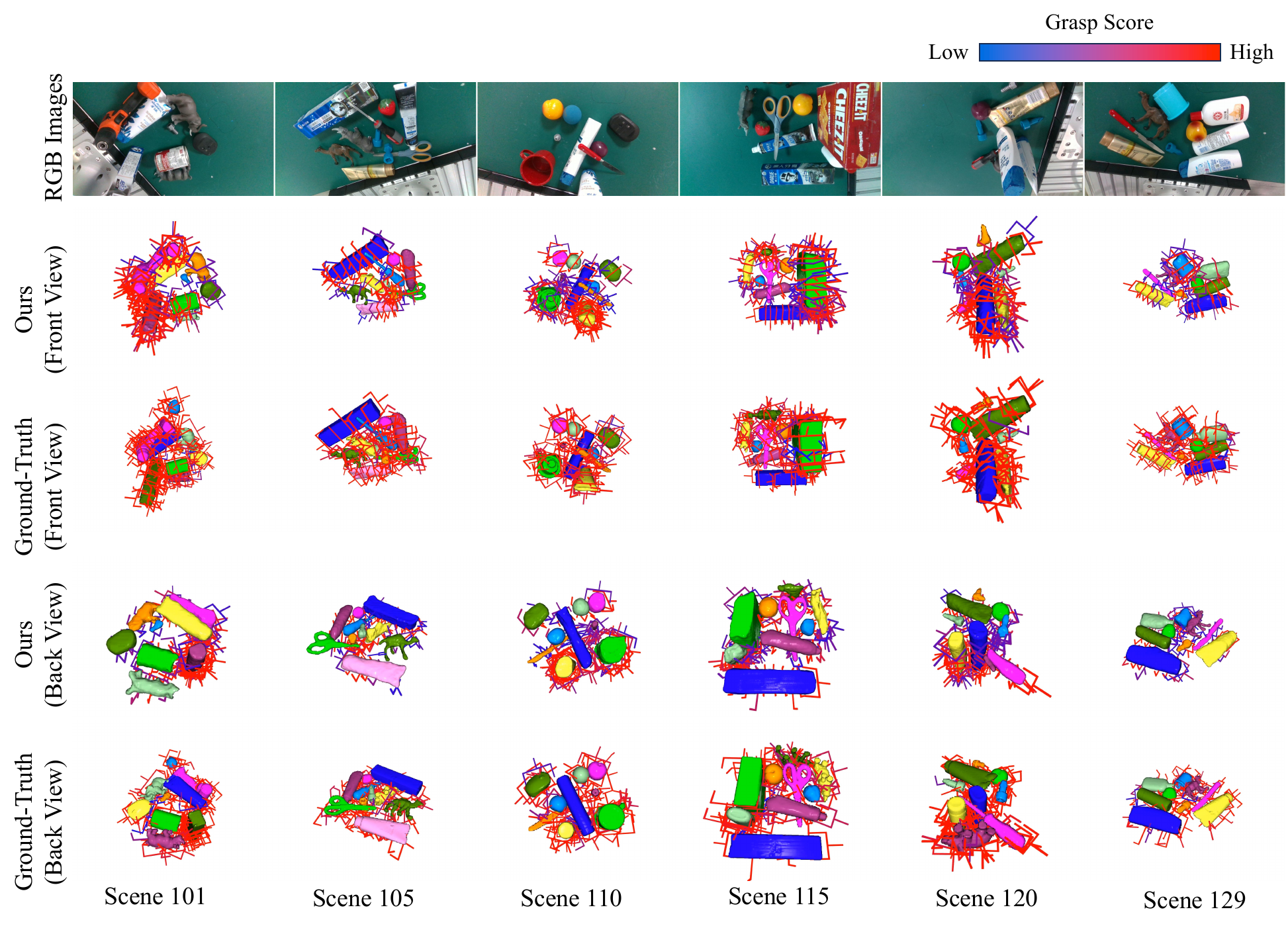}
    \vspace{-0.3cm}
    \caption{Results of grasp pose prediction on the seen split of the GraspNet-1B dataset. Grasp-NMS~\cite{fang2020graspnet} is applied to discard redundant grasps for better visibility of 3D reconstructions and grasp poses.}
    \label{fig:grasp_graspnet_seen}
\end{figure*}

\begin{figure*}[t]
    \centering
    \includegraphics[width=0.81\linewidth]{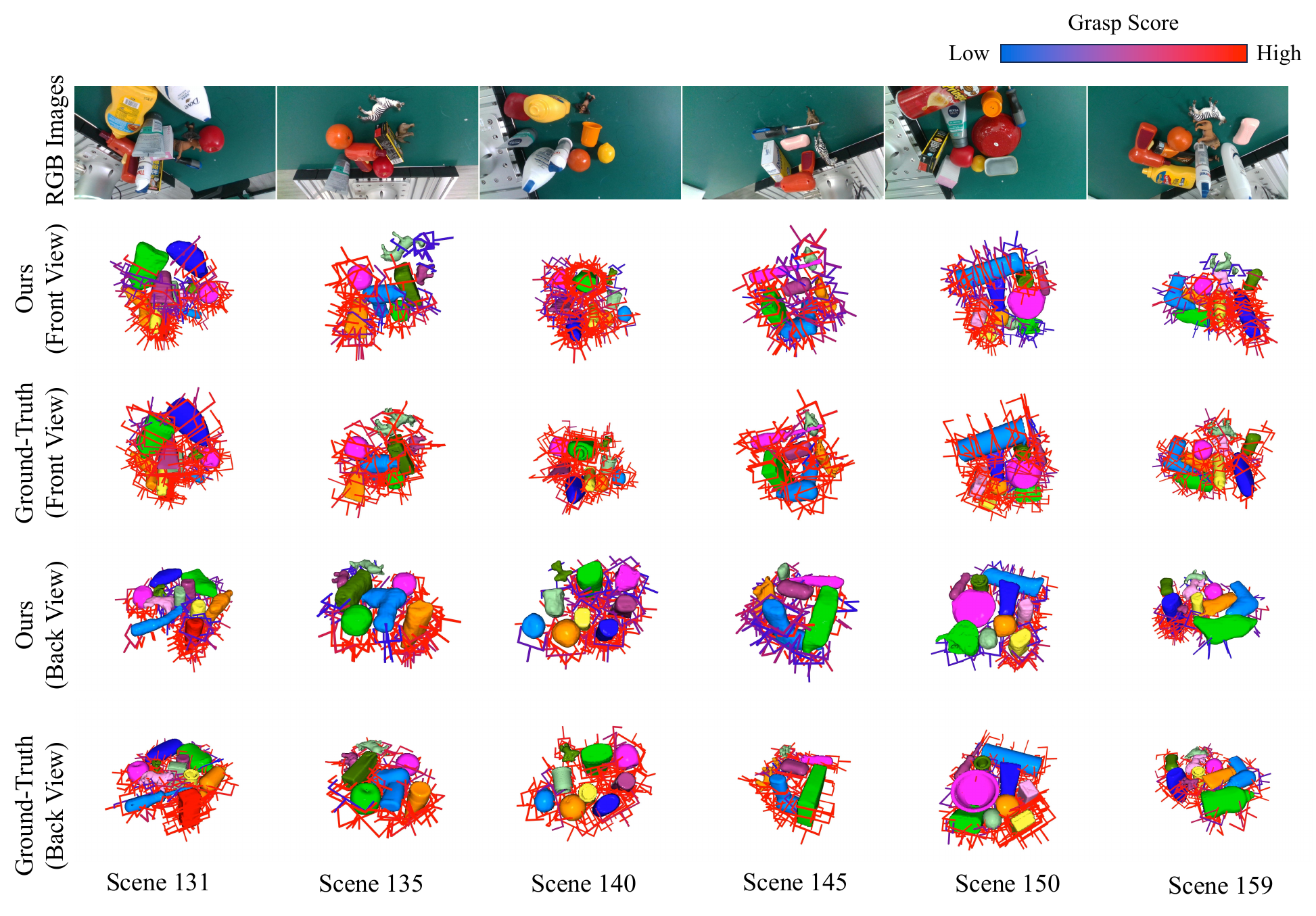}
    \vspace{-0.3cm}
    \caption{Results of grasp pose prediction on the similar split of the GraspNet-1B dataset. Grasp-NMS~\cite{fang2020graspnet} is applied to discard redundant grasps for better visibility of 3D reconstructions and grasp poses.}
    \label{fig:grasp_graspnet_similar}
\end{figure*}

\begin{figure*}[t]
    \centering
    \includegraphics[width=0.81\linewidth]{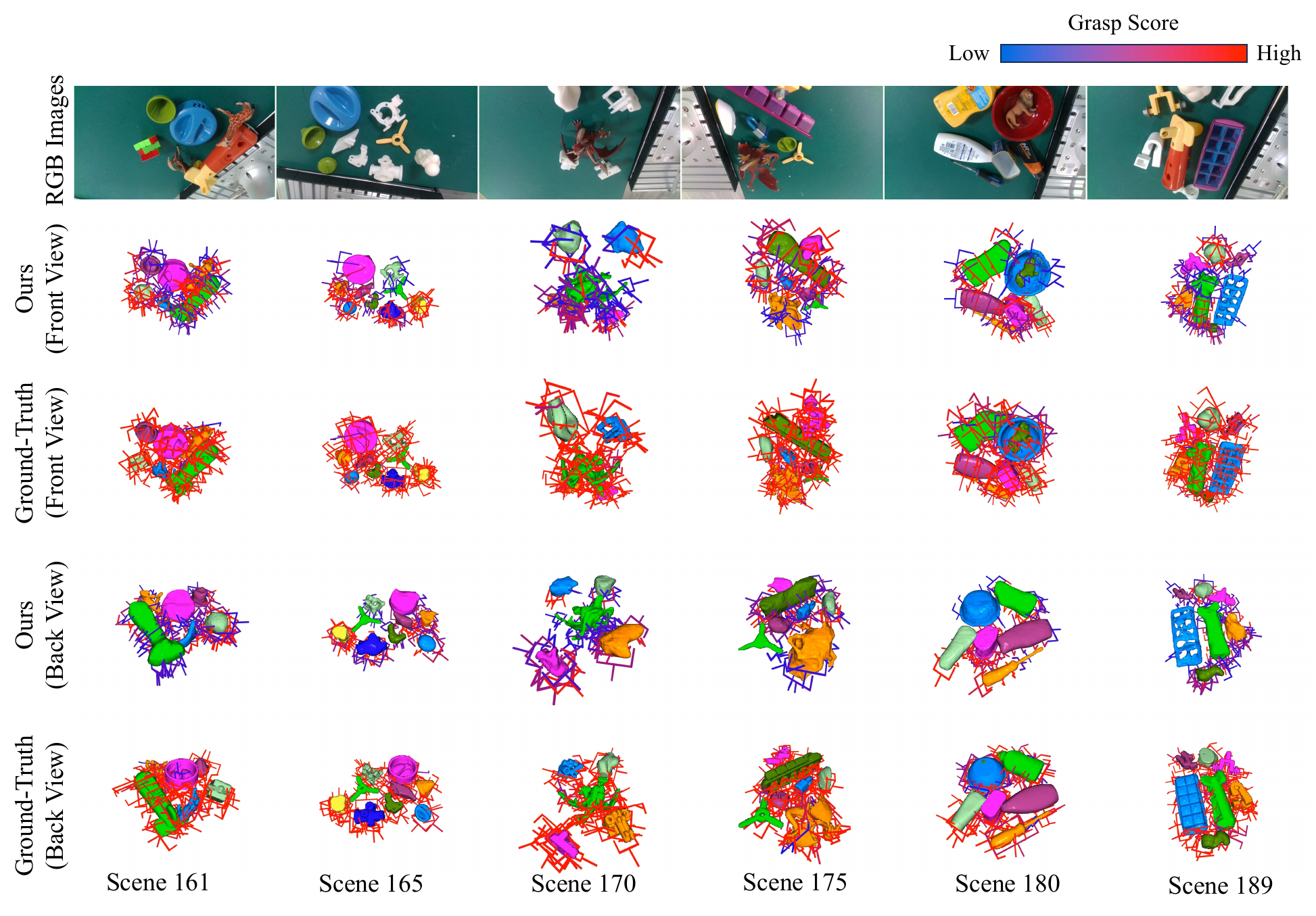}
    \vspace{-0.3cm}
    \caption{Results of grasp pose prediction on the novel split of the GraspNet-1B dataset. Grasp-NMS~\cite{fang2020graspnet} is applied to discard redundant grasps for better visibility of 3D reconstructions and grasp poses.}
    \label{fig:grasp_graspnet_novel}
\end{figure*}

We showcase qualitative results for 3D reconstruction (\Cref{fig:recon_reocs_easy,fig:recon_reocs_normal,fig:recon_reocs_hard}) and grasp pose prediction (\Cref{fig:grasp_graspnet_seen,fig:grasp_graspnet_similar,fig:grasp_graspnet_novel}). These examples are randomly sampled from each split or scene, highlighting ZeroGrasp's ability to handle a wide range of target objects. For turntable visualizations and real-robot evaluations, please refer to \href{https://sh8.io/\#/zerograsp}{our website}.

\subsection{Runtime Analysis}
The inference speed on NVIDIA A100 is $212$ ms with GPU memory usage below 8GB. Average runtimes for GraspNet, GSNet, and Ma \etal are $121$ ms, $98$ ms, and $238$ ms, respectively. Despite additionally reconstructing 3D objects, our runtime remains near real-time and our method still achieves the best AP metric.

\subsection{Analysis on Segmentation Masks}
We tested ZeroGrasp with ground truth masks but observed only a marginal improvement of +0.45 of AP on average in the GraspNet-1B benchmark. We perform an additional experiment where ZeroGrasp uses a foreground mask instead of an instance mask sand performs reconstruction and grasp pose prediction at the scene level. In practice, the F-1 score for reconstruction on the hard split of the ReOcS dataset drops to 80.23, which is 0.63 lower, primarily due to artifacts merging multiple objects. AP scores on GraspNet-1B show minimal change on average, with 69.91 (-0.62), 62.37 (-0.14), and 27.21 (+0.75) for the seen, similar, and novel splits, respectively. This suggests that foreground masks can replace instance masks, but at the cost of reduced reconstruction quality and 3D instance segmentation.

\end{document}

%% file: preamble.tex
\usepackage[dvipsnames]{xcolor}